\documentclass[10pt,twocolumn,letterpaper]{article}

\usepackage{enumitem,amssymb}
\usepackage{pifont}
\usepackage{multirow}
\usepackage[normalem]{ulem}
\useunder{\uline}{\ul}{}
\usepackage{hhline}
\usepackage{color}
\usepackage{caption}
\usepackage{graphicx}
\usepackage{multirow}
\usepackage{booktabs}
\usepackage{arydshln}
\usepackage{dblfloatfix}
\usepackage{subcaption}
\usepackage[table]{xcolor}
\usepackage{makecell}

\newcommand{\x}{\boldsymbol{x}}

\newcommand{\gaussian}{\mathcal{N}(\mathbf{0}, \mathbf{I})}
\newcommand\numberthis{\addtocounter{equation}{1}\tag{\theequation}}

\usepackage{iccv}
\usepackage{times}
\usepackage{epsfig}
\usepackage{graphicx}
\usepackage{amsmath}
\usepackage{amssymb}
\usepackage{subfiles}
\usepackage{float}
\usepackage{booktabs}
\usepackage[pagebackref=true,breaklinks=true,colorlinks,bookmarks=false]{hyperref}
\iccvfinalcopy
\newcommand{\acro}{GradPaint}

\begin{document}
\title{\acro{}: Gradient-Guided Inpainting with Diffusion Models}

\author{Asya Grechka, Guillaume Couairon, Matthieu Cord\\
Sorbonne Université\\
{\tt\small \{asya.grechka, guillaume.couairon, matthieu.cord\}@isir.upmc.fr}
}
\maketitle
\begin{abstract}

Denoising Diffusion Probabilistic Models (DDPMs) have recently achieved remarkable results in conditional and unconditional image generation. The pre-trained models can be adapted without further training to different downstream tasks, by guiding their iterative denoising process at inference time to satisfy additional constraints. For the specific task of image inpainting, the current guiding mechanism relies on copying-and-pasting the known regions from the input image at each denoising step. However, diffusion models are strongly conditioned by the initial random noise, and therefore struggle to harmonize predictions inside the inpainting mask with the real parts of the input image, often producing results with unnatural artifacts.

Our method, dubbed GradPaint, steers the generation towards a globally coherent image. At each step in the denoising process, we leverage the model's ``denoised image estimation" by calculating a custom loss measuring its coherence with the masked input image. Our guiding mechanism uses the gradient obtained from backpropagating this loss through the diffusion model itself. GradPaint generalizes well to diffusion models trained on various datasets, improving upon current state-of-the-art supervised and unsupervised methods. Our code will be made available upon publication.

\end{abstract}
\section{Introduction}

Inpainting consists in generating a missing part of a given image, given a binary mask indicating where the generation should take place. It is a fundamental task in computer vision, having obvious implications for image editing, image restoration, object removal, and so on. Currently, state-of-the-art methods are generally based on Generative Adversarial Networks (GANs) \cite{lama, zhao2021comodgan}, and consist in explicitly training a model to reconstruct an image using self-generated masks. Although these methods often achieve reasonable results with standard metrics, visual results tend to have obvious, unrealistic artifacts. Moreover, training these models is accompanied with the difficulties of training instability inherent with GANs as well as limitations on the diversity of the dataset distribution. 

Denoising diffusion probabilistic models (DDPMs) have recently gained massive attention, achieving high-resolution, photo-realistic and diverse image generation \cite{dalle2, imagen, latentdiffusion, latentdiffusion2, guided-diffusion, glide}. In terms of image generation, these models are on par or better than GANs even for constrained datasets like faces \cite{latentdiffusion}; and largely surpass them for diverse datasets like ImageNet \cite{glide, latentdiffusion}. Furthermore, recent models trained on large-scale datasets \cite{imagen, dalle2, glide, makeascene, latentdiffusion} have given rise to high-quality and flexible text-conditioned image generation, allowing users to generate astonishingly imaginative or artistic high-resolution images \cite{artcomp}. It is thus highly enticing to be able to use these pretrained models directly for downstream tasks, rather than re-training a new model from scratch. Here, we focus on the particular downstream task of inpainting.

There has been limited work in using pre-trained diffusion models for this task, and the typical approach \cite{repaint, meng2022sdedit, glide} is to guide the generative model by replacing values of the intermediate noise map with noised pixels of the input image outside the inpainting mask, based on the hope that the denoising process inside the inpainting mask will progressively be biased towards image parts that blend naturally with the known surrounding context.
However, this strategy often produces unsatisfying results, which we believe is due to the diffusion model being strongly conditioned on the initial noise map \cite{optimaltransport}, therefore having difficulties harmonizing the generation when the initial random latent map is too mismatched with the input image.

In this paper, we propose a new strategy for guiding pre-trained diffusion models
to better perform inpainting tasks. Our method, dubbed GradPaint, is optimizing the diffusion process by better harmonizing generated content inside the inpainting mask. This guides the generation at every single step of the denoising process towards a more harmonized final image.  Our method aims to minimize or even eliminate all the artifacts and inconsistencies that generally persist on the images due to the masked regions.
We propose a training-free algorithm which is advantageous because (i) there is no need to train a inpainting-specialized model whenever a new model is available, and (ii) training-based methods must chose a mask distribution to train on, to which training-free methods are agnostic. We perform an extensive evaluation on various datasets, including  CelebA-HQ\cite{celebahq}, FFHQ\cite{ffhq}, ImageNet\cite{imagenet}, Places2\cite{zhou2017places}, and COCO\cite{cocodataset}. 

Our main contributions can be summed up as:

\begin{itemize}
    \item We propose a novel training-free algorithm to the denoising scheduling of diffusion models for the specific task of inpainting. We improve this inpainting mechanism with the explicit goal of harmonizing the generated parts with the current context. Specifically, we use a custom \textit{alignment loss} and leverage the intrinsic nature of diffusion models through which we back-propagate and calculate a gradient to optimize our loss. 

    \item We show that our method generalizes well to a variety of datasets and pre-trained models, including latent-diffusion models. We show that our method improves baseline methods and is even on par with equivalent models trained specifically for the task of inpainting.
\end{itemize}
\section{Related Work}

\subsection{Inpainting}

Historically, inpainting was aimed at recovering small corruption errors in images and was addressed with matching or ``borrowing" local color and texture around the masked region \cite{poisson, patch_based}. Evaluation consisted in calculating a distance metric with respect to the unmasked image. More recently, generative models have become capable of synthesizing realistic and diverse images, allowing the use of much larger masks when inpainting images. Generative models thus have more freedom to ``imagine" a wide range of possibilities much different from the reference image, which is satisfactory (and oftentimes desired) so long as the resulting output looks realistic. 

In recent years, inpainting has been primarily addressed with training deep encoder-decoder convolutional networks from scratch, often using a GAN\cite{goodfellowgans} loss to encourage plausibility. Most recent work consists in improving the typical convolutional architecture in the encoder and/or decoder to better leverage structural or textural information from the surrounding regions \cite{lama, hong2019deep, yu2020region, hukkelaas2020image, yang2020learning, zhu2021image, liu2018image, ma2022regionwise, zheng2022cm}. \cite{li2020recurrent} proposes a progressive inpainting scheme which iteratively fills in the mask by using surrounding information in the deep feature space.  \cite{xiong2019foreground, liao2020guidance} propose a framework to locate and leverage semantic information.

In another line of work similar to ours, image completion is effectuated with the help of existing priors not specifically trained for the task. \cite{ulyanov2018deep} trains a randomly initialized convolutional network to generate the input image, stopping training before overfitting occurs. \cite{psp, zhao2021comodgan, glean} utilize powerful pre-trained decoders like StyleGAN2\cite{stylegan2} and only train encoders to map the input image into the latent space of the decoder, which can produce more realistic results if the input image fits well to the distrubtion of the pre-trained decoder.

\begin{figure*}[ht]
  \centering
    \includegraphics[width=\textwidth]{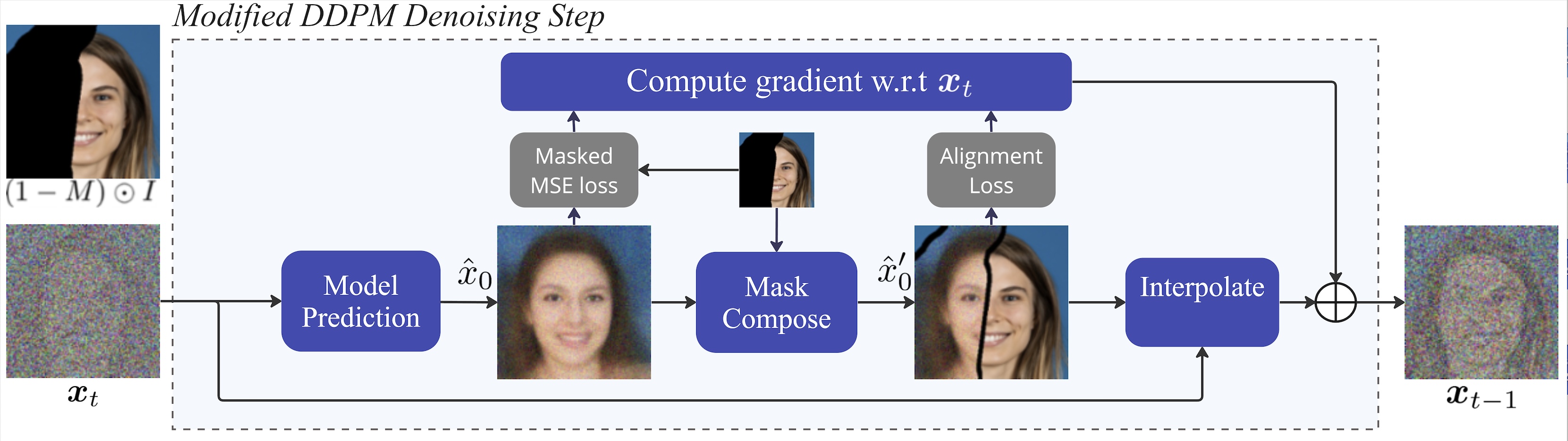}
    \caption{GradPaint method overview. We propose to modify one step of the DDPM denoising process with a gradient descent update on $x_t$ to better match the masked input image, in turn producing a better matched noise map $x_{t-1}$ for the next step. This improvement in the DDPM noise prediction thus allows for better fitting intermediate noise map predictions $x_t$ earlier in the DDPM denoising process, which ultimately produces a successful final inpainted image $x_0$.}
    \label{fig:method}
\end{figure*}

\subsection{Diffusion models}

Diffusion models are becoming state-of the art methods for generation tasks on many modalities, like images, videos, speech and text. Their excellent scaling behavior makes them a model of choice for training on large and diverse data, compared to GANs which still suffer from mode collapse and training instabilities. They can also be conditioned on various input data: for the specific task of inpainting, the input image and mask can be given as additional input to train a conditional diffusion model specialized on the inpainting task, as done in \cite{saharia2022palette}.

However, due to the computational cost of training generative models, it is appealing to find adaptation algorithms for downstream tasks without fine-tuning, especially for the task of inpainting which bears a lot of similarities with the unconditional generation task. \cite{latentdiffusion, glide, song2021scorebased} propose to adapt pre-trained diffusion models to inpainting by injecting a guiding mechanism in the generative process, a strategy which we build upon in this paper. \cite{repaint} also proposes to take advantage of pre-trained diffusion models with cycles of denoising and renoising operations, which we found computationally very expensive. Finally, in a parallel line of work most similar to ours, \cite{mcg} similarly propose to guide the generation using the gradient of a ``manifold constraint", but they do not use a custom loss nor do they apply optimization to the entirety of the intermediate noise maps.

\section{\acro{} Method}

\subsection{Background}
\label{background}

Denoising diffusion probabilistic models \cite{ho2020denoising} is a class of generative models trained with the following image denoising objective:

\begin{equation}
    \mathcal{L} = \displaystyle \mathbb{E}_{\x_0, t, \epsilon} \Vert \epsilon - \epsilon_\theta(\x_t, t) \Vert_2^2,
\end{equation}
where $\epsilon_\theta$ is a noise estimator network trained to predict the noise $\epsilon \sim \gaussian$  mixed with an input image $\x_0$ in the following way: $\x_t = \sqrt{\alpha_t} \x_0 +  \sqrt{1 - \alpha_t} \epsilon$. This training is performed for different values of the mixing coefficient $\alpha_t$, monotonically decreasing from $\alpha_0 = 1$ (no noise) to $\alpha_T \simeq 0$ (almost pure noise) for a large integer $T$.

At inference time, a new sample from the training distribution can be obtained by starting from random Gaussian noise $\mathbf{x}_T \sim \mathcal{N}(\mathbf{0}, \mathbf{I})$, and iteratively refining it with the noise estimator network with the following equations, called \textit{DDPM sampling equations} \cite{ho2020denoising}:

$x$ and $\x$
\begin{align*}
\hat{\x}_0 & = \frac{1}{\sqrt{\alpha_t}}(\x_t - \sqrt{1 - \alpha_t} \cdot \epsilon_\theta(\x_t, t)), \numberthis \label{eq:ddpm1}\\
\x_{t-1} & = \frac{(\alpha_{t-1}-\alpha_t) \sqrt{\alpha_{t-1}}}{\alpha_{t-1}(1 - \alpha_t)} \hat{\x}_0 + \frac{(1-\alpha_{t-1})\sqrt{\alpha_t}}{(1 - \alpha_t)\sqrt{\alpha_{t-1}}} \x_t + \sigma \boldsymbol{z},
\end{align*}
where $t$ goes from $T$ to $0$, 
$\sigma_t$ is a variance parameter, and $z \sim \gaussian$.

This iterative refinement can be ``guided" to impose constraints on the generated sample $\x_0$. In the case of inpainting, the aim is that the generated image exactly matches the input image outside a given inpainting region. The variable $\hat{\x}_0$, available at each timestep, represents the model's current estimation of what the denoised image will look at the end. For instance, \cite{glide} applies a maskwise correction on $\hat{\x}_0$ at each timestep:
\begin{equation}
\hat{\x}_0' = M \odot \hat{\x}_0 + (1 - M) \odot I,
\end{equation}
where $I$ is the input image and $M$ is a binary image mask equal to 1 in the image regions that must be inpainted, 0 otherwise. The update rule for $\x_{t-1}$ is then adapted to use $\hat{\x}_0'$ instead of $\hat{\x}_0$ in \autoref{eq:ddpm1}. This correction progressively biases the diffusion model to exactly match $I$ outside the inpainting mask $M$. In the remaining of the paper, we refer to this method as \textit{combine-image} since it combines the images $\hat{\x}_0$ and $I$ before interpolating with $\x_t$.

Alternatively, \cite{song2021scorebased, latentdiffusion, repaint} propose to directly correct $\x_{t-1}$ by replacing regions outside $M$ with the noised regions of the input image $I$:
\begin{equation}
\x_{t-1}' = M \odot \x_{t-1} + (1-M) \odot (\sqrt{\alpha_{t-1}} I +  \sqrt{1 - \alpha_{t-1}} \epsilon),
\end{equation}
where $\epsilon \sim \gaussian$ is resampled at each step. This $\x_{t-1}'$ is then used as input for the next denoising step instead of $\x_{t-1}$. We will refer to this method as \textit{combine-noisy} since it combines $\x_{t-1}$ inside the mask with ground truth (noised) pixel values outside the mask.

\subsection{\acro{} framework}

Our strategy is built upon the \textit{combine-image} zero-shot inpainting method presented in \S\ref{background}. Our key observation is that the most asthetically-pleasing inpainting results are obtained when the collage $M \odot \hat{x}_0 + (1-M) \odot I$ is coherent right from the beginning of the generation process. When this is not the case, there is a mismatch between the model's estimation in the inpainting region and the known regions of input image $I$. This mismatch is generally present from the beginning and is not fully corrected during the denoising generation process. 

To enforce harmonization between the inpainted region and known regions of the input image, we introduce the \textit{GradPaint update}. An overview of our method is presented in Fig.~\ref{fig:method}.
At each denoising step, the variable $\x_t$ is updated so that (i) $\hat{x}_0$ matches well known regions of $I$ outside the mask; and (ii) the collage $M \odot \hat{x}_0 + (1-M) \odot I$ does not present any discontinuity due to the copy-paste operation. This update consists in a one-step gradient descent update from two loss terms corresponding to the two objectives aforementioned. 

Given a binary mask $M \in \mathbb{R}^{n \times n}$ and $\odot$ denoting the element-wise product, we define our losses as follows:

\noindent \textbf{Masked MSE loss.} The first loss term is a mean squared error term outside the inpainting mask $(1 - M)$, taking as reference known regions of the input image:
\begin{equation}
    \mathcal{L}_{mse}(I_1, I_2, M) = \frac{1}{n^2}\Vert I_1 \odot (1 - M) - I_2 \odot (1 - M) \Vert_2^2. 
\end{equation}

\noindent \textbf{Alignment loss.} The ``alignment loss" $al(I, M)$ measures the smoothness of image $I$ on the boundaries of the inpainting mask $M$. It is defined as follows:
\vspace*{-.1cm}\begin{equation}
    al(I, M) \hspace{-0.1cm}=  \hspace{-0.1cm} \frac{1}{n^2}\Vert D_x I \odot D_x (1 - M) +D_y I \odot D_y (1 - M) \Vert_2^2, 
\end{equation}
where $D_x$ and $D_y$ are the normalized image gradients:

\begingroup\makeatletter\def\f@size{9}\check@mathfonts
\def\maketag@@@#1{\hbox{\m@th\large\normalfont#1}}%
\begin{align*}
 \begin{bmatrix}
D_x I \\
\vspace{-0.2cm}\\
D_y I \\
\end{bmatrix}_{(i, j)}
&= 
\begin{cases}
\dfrac{\nabla I_{(i, j)}}{||\nabla I_{(i, j)}||_2} , & { \text{if } ||\nabla I||_{(i, j)} > 0} \vspace{0.1cm} \numberthis \\
[0 \quad 0]^T
, & \text{else}
\end{cases} \\
\end{align*}\endgroup

\noindent with $\nabla I = [\partial_x I \; \partial_y I]^T$ is the vector of gradients of $I$ in the $x$ and $y$ directions respectively. 
When we minimize this loss, we aim to achieve the smoothest transition possible in the image $I$ along the direction where $M$ changes values. 
Since this loss $al(I, M)$  is defined for an image with only one color channel, we simply define the total alignment loss $\mathcal{L}_{al}$ as the average loss over the three color channels for a regular RGB image.

\noindent \textbf{\acro ~Update.} Our total loss is defined as:

\begin{equation}
\mathcal{L} = \mathcal{L}_{mse} + \lambda_{al} \mathcal{L}_{al},
\end{equation}
with $\lambda_{al}$ being a hyperparameter controlling the relative strength of the alignment loss compared to the MSE loss.

At each step in the denoising process, we compute $\x_{t-1}$ as a function of $\x_t$ as in the \textit{combine-image} method. In between each step, we update the variable $\x_{t-1}$ with the normalized gradient of our total loss:

\begin{equation}
\x_{t-1}' = \x_{t-1} - \alpha \frac{\nabla_{\x_t}\mathcal{L}(x_0, \hat{x}_0, M)}{\Vert \nabla_{\x_t}\mathcal{L}(x_0, \hat{x}_0, M) \Vert_2},
\end{equation}
with $\alpha$ being a fixed learning rate.

Backpropagating through the diffusion model itself until variable $\x_t$ is a crucial element of our method. Since $\x_t$ is updated to produce a better estimation $\hat{\x}_0$ when processed by the diffusion model, this property will also transfer to $\x_{t-1}$ which is, at each step, very close to $\x_t$.

\subsection{Visualizations}

\begin{figure}[htbp]
  \centering
    \includegraphics[width=\linewidth]{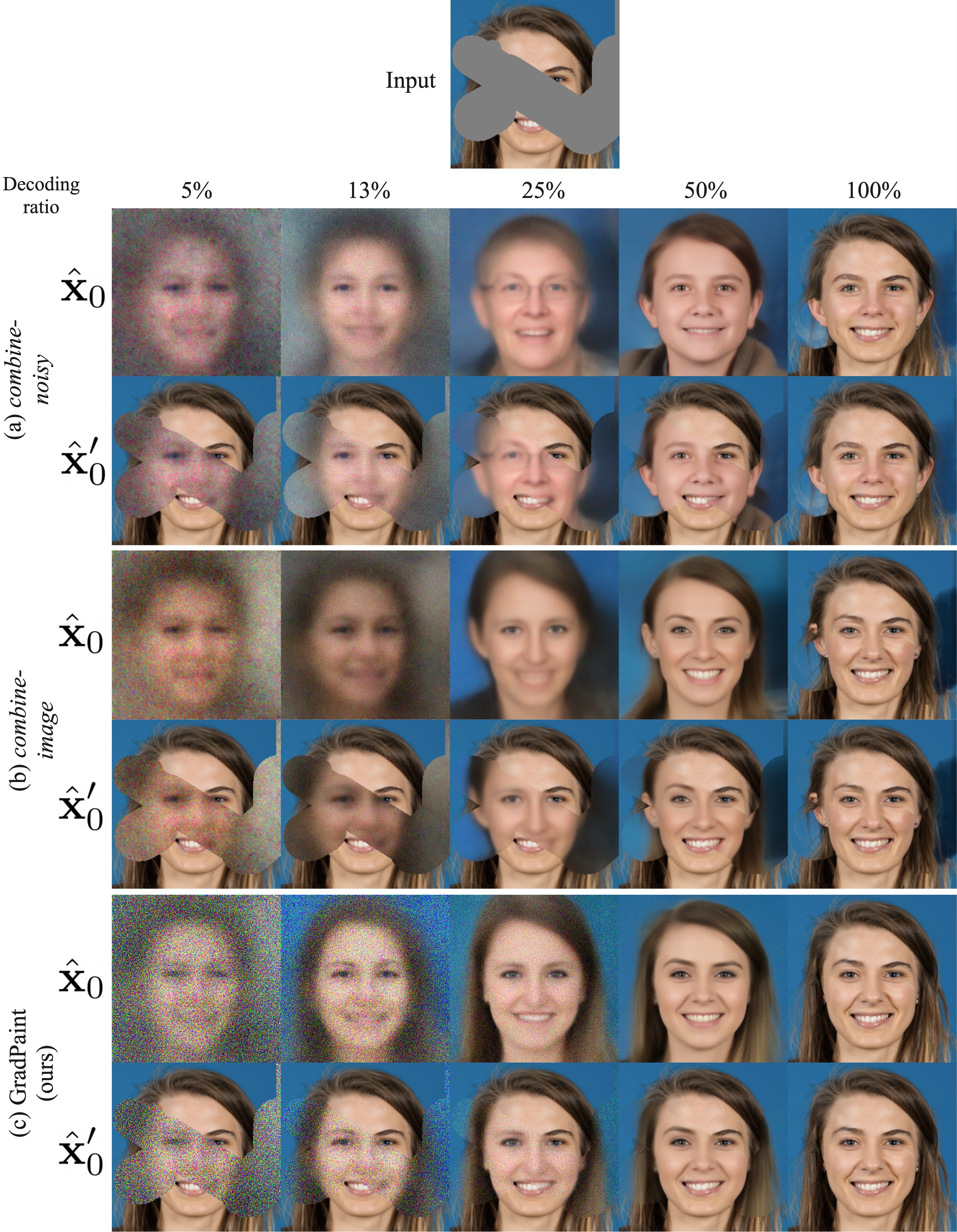}
    \caption{DDPM predictions at different stages (indicated in $\%$) of the denoising process. We compare two baselines (a) and (b) with \acro{} (the two last rows). \acro{} better harmonizes regions inside and outside the inpainting mask right from the beginning of the denoising process.}
    \label{fig:intuition}
\end{figure}

\noindent \textbf{Harmonization.} The effect of the GradPaint update is illustrated in Fig.~\ref{fig:intuition}, 
which shows the intermediate DDPM predictions for $\hat{\x}_0$ and $\hat{\x}_0'$ at various timesteps. We 
compare \acro{} with the \textit{combine-noisy} and \textit{combine-image} methods presented in \S\ref{background}, where all three methods share the same DDPM model, parameters and initial noise maps.
These baseline approaches require more steps to integrate the information from the input image, at which point it is often ``too late" to construct a harmonized image - misalignment between the generation and the input image can no longer be corrected. In contrast, for GradPaint, the optimization step on $\x_t$ quickly pushes the merged image $\hat{\x}_0'$ to harmonizes well with the masked input image $\x_0$, producing an inpainting result without alignment artifacts.

\begin{figure*}[htbp]
  \begin{subfigure}[b]{0.50\linewidth}
    \includegraphics[width=\linewidth]{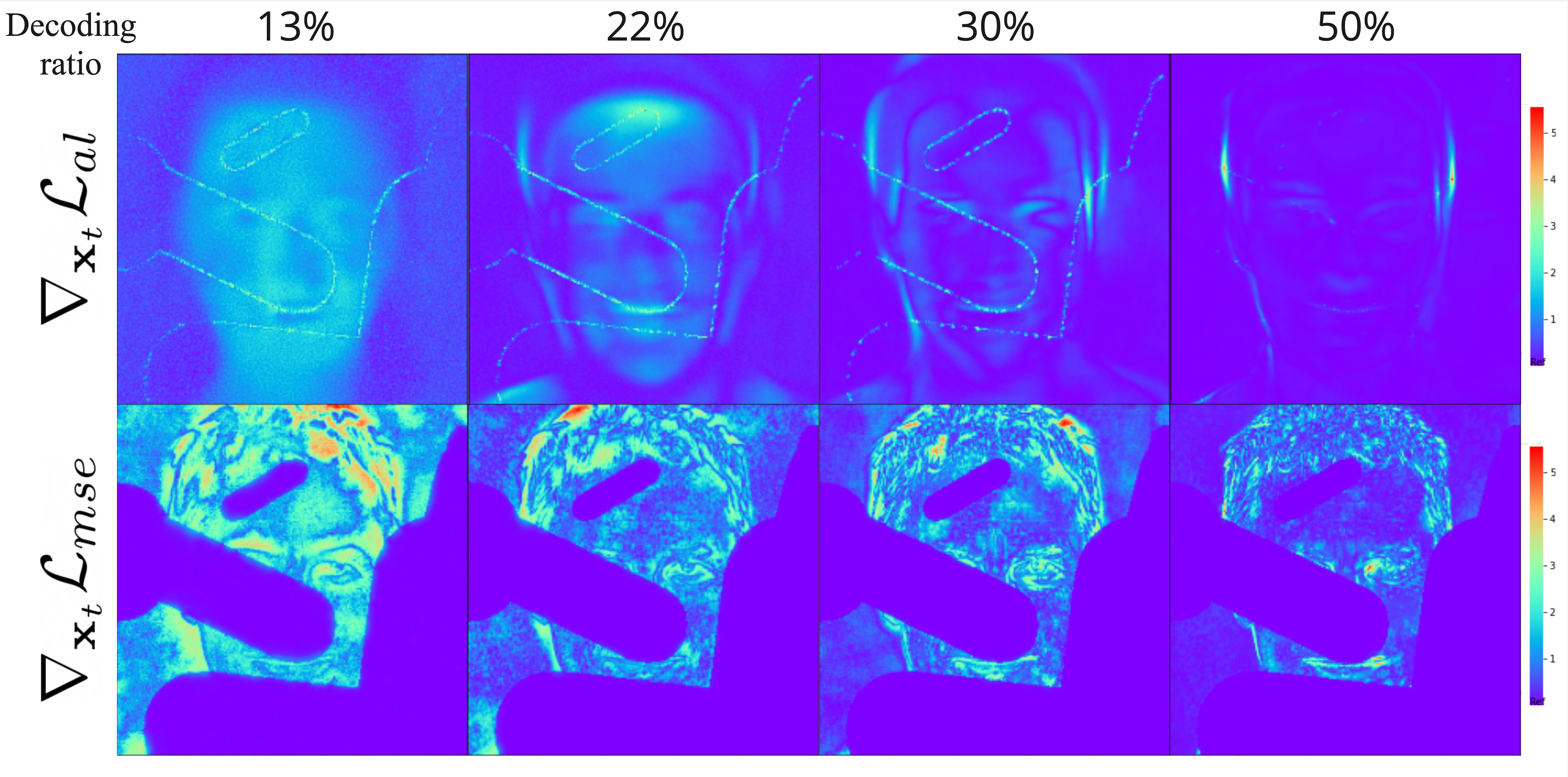}
    \caption{Gradient magnitude of different components of our losses with regards to $x_t$. The alignment loss has a concentrated effect at the border and a more global effect compared to the masked MSE loss, but dies out more quickly when it concentrates in a few local spots.}
    \label{fig:loss_intuition-grad}
  \end{subfigure}
  \hfill 
  \hspace{0.3cm}
  \begin{subfigure}[b]{0.45\linewidth}
  \centering
    \includegraphics[width=0.8\linewidth]{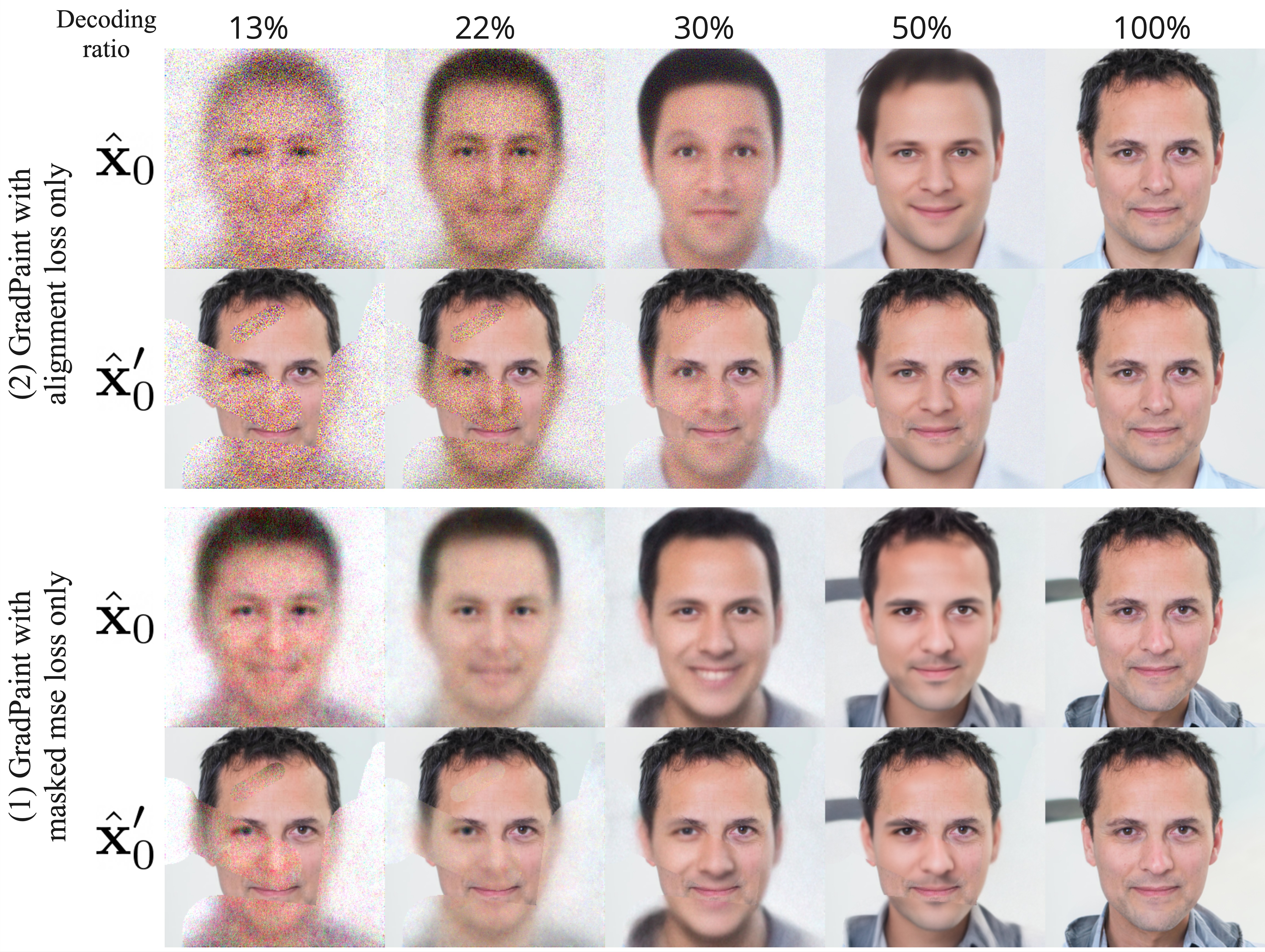}
    \caption{Intermediate DDPM predictions with GradPaint using separate components of our loss. The alignment loss encourages smooth and coherent transitions, as can be seen with the homologous background.}
    \label{fig:loss_intuition-int}
  \end{subfigure}
  \caption{Effect of separate components of our loss on the intermediate predictions of the DDPM model and their corresponding gradients. Noise maps are initialized identically.}
  \label{fig:loss_intuition}
\end{figure*}

\noindent \textbf{Gradient visualization.} The two separate components of our loss have different effects on $\nabla{\x_t}$, as we can see in Fig.~\ref{fig:loss_intuition-grad}.  While the gradient of the masked MSE loss remains active throughout the denoising process, the gradient of the alignment loss becomes obsolete about halfway-through, thereafter only concentrating in a few local points in $\x_t$. The gradient of the alignment loss has a concentrated effect on the borders of the mask, but also affects the entire noise map $\x_t$ globally, while the masked MSE loss has a much stronger effect in the unmasked region. The alignment loss encourages smoother and more gradual transitions in the final generation, as can be seen with the background in Fig.~\ref{fig:loss_intuition-int}.

\section{Evaluation Protocol}

\subsection{Pre-trained models and implementation details} 
We detail our setup for image-space diffusion models as well as latent-space diffusion models.
We provide a detailed list of the assets 
used in our work (datasets, code, and models) in
the Appendix in~\ref{sec:GradPaint assets}.

\paragraph{Experiments on image-space diffusion models}
We primarily use diffusion models from guided diffusion~\cite{dhariwal2021diffusion}, which operates on images of size $256\times256$. 
We use their pre-trained unconditional models (pre-trained on FFHQ, CelebaHQ, and Places2)  as well as their class-conditional model
trained on ImageNet. 

We use a default number of 100 steps for DDPM 
sampling; the loss is computed with $\lambda_{al} =400$ during the first 45 steps of decoding (and disabled afterwards following 
our observations shown in~\ref{fig:loss_intuition-grad}). The gradient is updated with a fixed learning rate of $0.005$.

\paragraph{Extension to latent diffusion models}

We also experiment with latent diffusion models~\cite{rombach2022high}.  We have observed that the latent spaces that we use have much
 less structure compared to real images, and that our alignment loss, whose role is to enforce smoothness on real images, 
 cannot fulfill this role in latent spaces. Therefore, for all experiments with latent diffusion models, we only experiment with 
 the masked MSE loss, which 
 naturally extends to latent spaces by considering the encoded input image as reference in our MSE loss.

Latent diffusion models also operate on $256\times256$ images, but images are
 edited in a latent space with spatial dimensions of $64\times64$.
 We use pre-trained unconditional latent diffusion models on CelebAHQ and FFHQ. We use the class-conditional 
 latent diffusion model pre-trained on ImageNet. Finally, for text-conditional models, we use Stable Diffusion pre-trained on the 
 public LAION-5B dataset~\cite{schuhmann2022laion}.

 We use a default number of 100 steps for DDPM sampling; the loss is computed with $\lambda_{mse} = 1$.
 The gradient is updated with a fixed learning rate of $0.005$.

\paragraph{Datasets}
We evaluate our algorithm on five datasets: FFHQ, CelebaHQ, ImageNet, Places2 and COCO.

Given an image, the aim is to perform inpainting inside a random mask generated with the mask generator from~\cite{lama}. 
We mainly evaluate on the difficult and more realistic \textit{thick} masks (additional results on \textit{thin} and \textit{medium} masks are provided in the appendix).  We create 5000 masked images for all experiments.

For both image-space and latent diffusion models, we evaluate the FFHQ pre-trained model on a subset of CelebAHQ  images. 
Inversely, we evaluate the CelebAHQ pre-trained model on a subset of FFHQ images. We evaluate the ImageNet pre-trained 
models on a subset ImageNet validation set and use the class label as conditioning. 

For the image-space diffusion model pre-trained on Places2, we use a subset of the Places2 validation set for evaluation. For the Stable Diffusion model pre-trained on LAION-5B, we use a subset of the COCO validation set and use the 
captions as conditioning text information for the diffusion model. 

\subsection{Metrics}

For a set of images inpainted with a given method, we compute two core metrics that encapsulate the challenges
 of inpainting: the LPIPS distance~\cite{zhanglpips2018} between the inpainted image and the (unmasked) input image which measures 
 the extent to which we correctly recover the masked regions, and the FID score~\cite{heusel2017gans} which measures the realism 
 of output images. The primary requirement is that inpainted images should look as natural as possible, hence having the smallest 
 possible FID score. For LPIPS distances, an inpainting result closer to the reference image is generally better, although realistic 
 images further away from the reference image can also be satisfactory, especially for large masks.

\subsection{Baselines}  We compute the best and worst possible LPIPS and FID scores with two trivial measures: 
the \textit{COPY} oracle measure, which simply copies the (unmasked) input image, gives an LPIPS score of $0$ and a lower bound on 
possible FID scores; and the \textit{GREYFILL} measure, which simply fills the region to be inpainted with uniform grey. 
We also add a \textit{Latent COPY} oracle for latent diffusion models which consists in simply auto-encoding the input image.
Without
 gradient-based optimization, our method is equivalent to the \textit{combine-image} baseline for image inpainting, which we evaluate 
 in our experiments along with its \textit{combine-noisy} variant. Apart from these three closely related methods, we compare against
  the following state-of-the-art inpainting methods:  LaMa~\cite{lama}, a GAN-based method trained for inpainting; 
  Palette~\cite{saharia2022palette}, also trained for inpainting but with diffusion models, RePaint~\cite{lugmayr2022repaint}, another
   training-free inpainting algorithm that is much more computationally expensive, and finally MCG~\cite{mcg}, a parallel line 
   of work to ours which is similarly training-free but with a different optimization scheme.

\section{Quantitative Evaluation}

\begin{table}[t]
\small
\begin{tabular}{|l|c|c|c|c|}
\hline
\multicolumn{1}{|r|}{Dataset} &
  \multicolumn{2}{c|}{FFHQ} &
  \multicolumn{2}{c|}{CelebaHQ} \\ \hline
\multicolumn{1}{|r|}{Metrics} &
  \multicolumn{1}{c|}{FID$\downarrow$} &
  \multicolumn{1}{c|}{LPIPS$\downarrow$} &
  \multicolumn{1}{c|}{FID$\downarrow$} &
  \multicolumn{1}{c|}{LPIPS$\downarrow$} \\
  \hline
  \rowcolor[gray]{0.7}
COPY (oracle) &
  4.29 & 0
  & 3.01 & 0 \\ \hline
GREYFILL &
  78.08 &
  0.257
  &
  96.41 &
  0.264
   \\ \Xhline{4\arrayrulewidth}
LaMa&
  6.27 &
  \textbf{0.076}
   &
  \multicolumn{1}{c|}{n/a} &
  \multicolumn{1}{c|}{n/a} \\ \hline
Palette &
  \multicolumn{1}{c|}{7.28} &
  \multicolumn{1}{c|}{0.096}
  &
  \multicolumn{1}{c|}{n/a} &
  \multicolumn{1}{c|}{n/a} \\ \Xhline{4\arrayrulewidth}
Repaint &
  \multicolumn{1}{c|}{9.09} &
  \multicolumn{1}{c|}{0.090} &
  \multicolumn{1}{c|}{8.44} &
  \multicolumn{1}{c|}{\textbf{0.078}}  \\ \hline
MCG &
  \multicolumn{1}{c|}{\underline{6.17}} &
  \multicolumn{1}{c|}{0.097}
  &
  \multicolumn{1}{c|}{6.67} &
  \multicolumn{1}{c|}{0.084} \\ \hline
\begin{tabular}[c]{@{}l@{}}\textit{combine-noisy}\end{tabular} &
  \multicolumn{1}{c|}{9.04} & 
  \multicolumn{1}{c|}{0.119}
   &
  \multicolumn{1}{c|}{9.89} & 
  \multicolumn{1}{c|}{0.103}
   \\ \hline
\begin{tabular}[c]{@{}l@{}}\textit{combine-image}\end{tabular} &
  \multicolumn{1}{c|}{7.30} &
  \multicolumn{1}{c|}{0.123}
   &
  \multicolumn{1}{c|}{\underline{5.83}} &
  \multicolumn{1}{c|}{0.110}\\ \hline
GradPaint (ours) &
  \multicolumn{1}{c|}{\textbf{5.65}} &
  \multicolumn{1}{c|}{{\underline{0.084}}}
   &
  \multicolumn{1}{c|}{\textbf{4.41}} &
  \multicolumn{1}{c|}{\textbf{0.077}} \\ \hline
\end{tabular}
 
\caption{Evaluation of various methods on FFHQ and CelebaHQ datasets. The COPY oracle and the GREYFILL measure are respectfully the lower and upper bounds for LPIPS and FID. LaMa and Palette are both training-based methods. RePaint, \emph{combine-noisy}, \emph{combine-image}, MCG and GradPaint are all training-free methods which all use the same model based on guided diffusion\cite{guided-diffusion}. Best score is shown \textbf{in bold} and second best \underline{underlined}. As we can see, our method produces the best results in terms of FID and is on-par with fully-supervised methods in terms of LPIPS.\vspace{1cm}}
\label{main_results}
\end{table}

\begin{table}
\begin{minipage}[t]{.60\linewidth}\vspace*{0pt}%
    \footnotesize
    \begin{tabular}{|c|l|l|}
\hline
Method                                                                  & FID $\downarrow$           & LPIPS $\downarrow$           \\ \hline
\textit{combine-noisy}                                                  & 10.33         & 0.1907          \\ \hline
\textit{combine-image}                                                  & 9.61          & 0.1797          \\ \hline
\begin{tabular}[c]{@{}c@{}}GradPaint \\ w/o alignment loss\end{tabular} & 8.12          & 0.1551          \\ \hline
GradPaint                                                               & \textbf{7.86} & \textbf{0.1486} \\ \hline
\end{tabular}
\end{minipage}
\hfill
\begin{minipage}[t]{.35\linewidth}
    \setlength{\abovecaptionskip}{0pt}%
    \caption{Detailed comparison on ImageNet pre-trained guided diffusion model with \emph{thick} masks.}%
    \label{tab:quantitative_ablation}
\end{minipage}
\end{table}

\paragraph{Image-space Diffusion Models}

Quantitative results on the FFHQ and CelebA datasets for image-space diffusion models are shown in Tab.~\ref{main_results}, 
where GradPaint is compared against 
available competing methods (FFHQ-pretrained checkpoints are not available for Palette and LAMA) as well as the \textit{combine} 
baselines. We present the results for the \emph{thick} mask setting, as this is the most interesting setting for practical applications. Results for other mask sizes are presented in the appendix. The benefit of our gradient update is visible when comparing to \textit{combine-image}
(same as ours without gradient updates): On FFHQ, the FID score is reduced from 7.30 to 5.65, a significant improvement given that 
the minimum obtainable FID score is 4.29 on 5000 images (\textit{COPY} oracle measure). Results on both datasets show similar gains. 
When comparing with competing methods on FFHQ, GradPaint obtains the state-of-the art FID score, outperforming methods specialized in 
inpainting (Palette, LAMA) as well as the training-free algorithms Repaint and MCG based on the same diffusion model as ours. LaMa 
obtains slightly better LPIPS scores but requires an inpainting-specific training (compared to simply using a pre-trained generative 
model). Moreover, LaMa, unlike all other methods, 
has access at train time to the mask distribution that we use for testing.

We validate different components of our method with the ImageNet~\cite{deng2009imagenet} dataset and  guided diffusion model, summarized in 
Tab.~\ref{tab:quantitative_ablation}. This more difficult dataset was chosen to better analyze our different components as well as 
validate our method on class-conditioned diffusion models, where generation could be biased by the class. Our full method, and the 
alignment loss in particular, improves reconstruction and realism of generated images.

\noindent \paragraph{Latent Diffusion Models}

 Results on Latent Diffusion Models are presented in Tab.~\ref{latent_table}. As we can see, the latent space allows for very good image 
 reconstruction (small LPIPS scores), so it is not a real limitation and GradPaint (latent) is still able to outperform competing 
 methods (FID 5.97 on FFHQ \emph{thick} masks). Overall, we observe large and consistent gains on all three datasets ImageNet, 
 COCO and FFHQ datasets over the reference inpainting methods, for both FID and LPIPS.

\begin{table}[ht]
\scriptsize
\begin{tabular}{|r|cc|cc|cc|cc|cc|}
\hline
Dataset &
  \multicolumn{2}{c|}{ImageNet} &
  \multicolumn{2}{c|}{COCO} &
  \multicolumn{2}{c|}{FFHQ}  \\ \hline
  
\multicolumn{1}{|c|}{} &
  \multicolumn{1}{c|}{FID} &
  LPIPS  &
  \multicolumn{1}{c|}{FID} &
  LPIPS &
  \multicolumn{1}{c|}{FID} &
  LPIPS \\ \hline
  \rowcolor[gray]{0.7}
COPY (o.) &
  \multicolumn{1}{c|}{12.27} &
  0.0 &
  \multicolumn{1}{c|}{7.29} &
  0.0 &
  \multicolumn{1}{c|}{4.29} &
  0.0
  \\ \hline
  \rowcolor[gray]{0.7}
Lat. COPY (o.) &
  \multicolumn{1}{c|}{12.00} &
  0.034 &
  \multicolumn{1}{c|}{7.71} &
  0.041 &
  \multicolumn{1}{c|}{4.98} &
  0.018
  \\ \hline
GREYFILL &
  \multicolumn{1}{c|}{34.51} &
  0.269 &
  \multicolumn{1}{c|}{29.97} &
  0.264 &
  \multicolumn{1}{c|}{77.43} &
  0.257 \\ \Xhline{4\arrayrulewidth}
\begin{tabular}[c]{@{}r@{}}\textit{combine-noisy}\end{tabular} &
  \multicolumn{1}{c|}{17.17} &
  0.195 &
  \multicolumn{1}{c|}{11.12} &
  0.241  &
  \multicolumn{1}{c|}{8.73} &
  0.132 
  \\ \hline
\begin{tabular}[c]{@{}r@{}}\textit{combine-image}\end{tabular} &
  \multicolumn{1}{c|}{17.37} &
  0.207 &
  \multicolumn{1}{c|}{12.68} &
  0.257 &
  \multicolumn{1}{c|}{6.832} &
  0.127
  \\ \hline
\begin{tabular}[c]{@{}r@{}}GradPaint\\ (ours)\end{tabular} &
  \multicolumn{1}{c|}{\textbf{14.62}} &
  \textbf{0.163} &
  \multicolumn{1}{c|}{\textbf{9.43}} &
  \textbf{0.216} &
  \multicolumn{1}{c|}{\textbf{5.97}} &
  \textbf{0.111}\\ \hline
\end{tabular}
\caption{Evaluation of pre-trained latent diffusion models with \emph{thick} masks. For all values, lower is better. The COPY oracle measures the metrics on the ground-truth images, and the Latent COPY oracle does the same for autoencoded ground-truth images. As we can see, our modification for latent diffusion models yields significant improvements on all datasets.}
\label{latent_table} 
\end{table}

\begin{figure}[ht]
  \centering
    \includegraphics[width=0.92\linewidth]{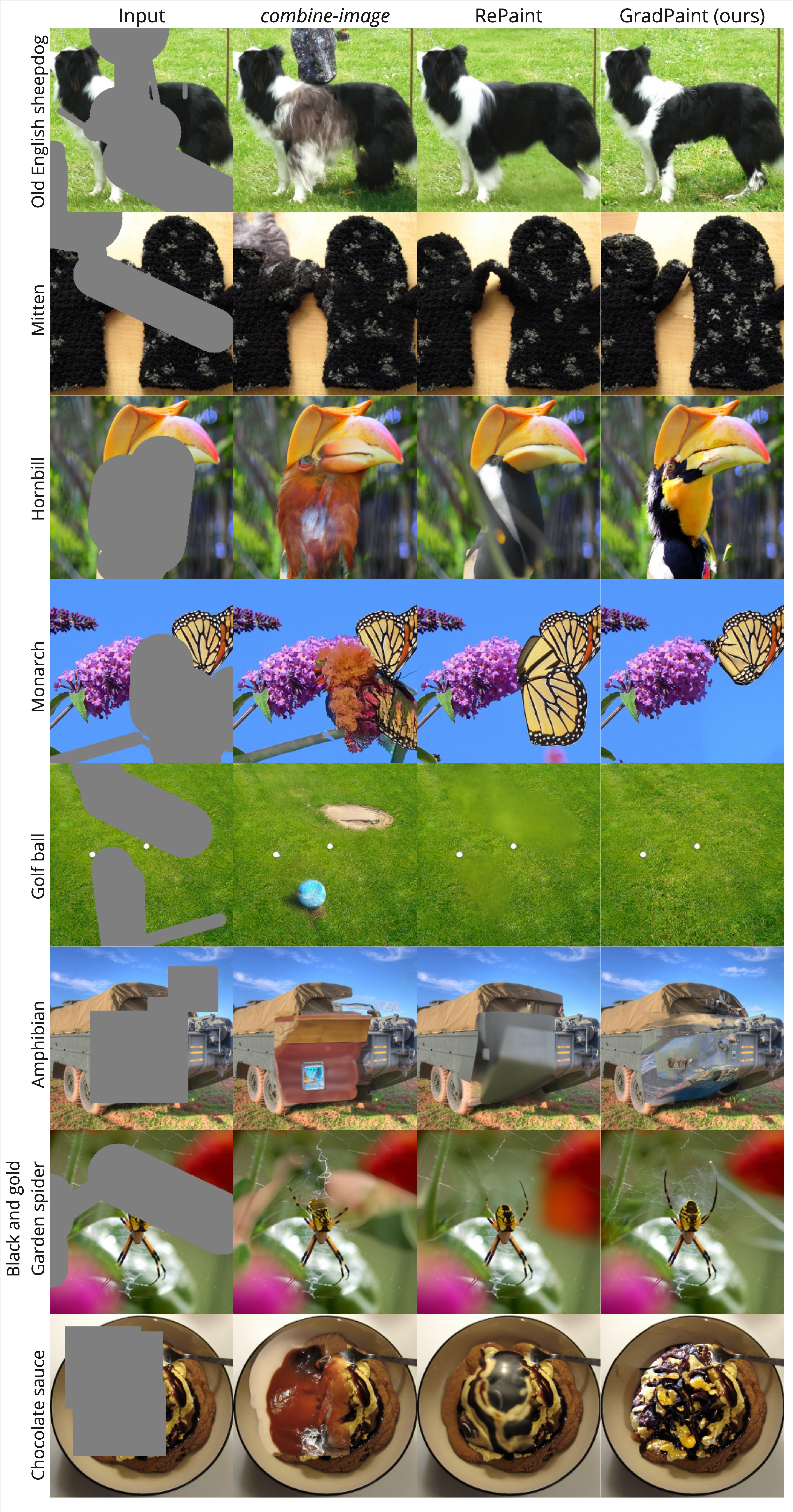}
    \caption{Inpainting results on select images from ImageNet. The \emph{combine-image} baseline produces unharmonized results and struggles to take the context into account. Our method produces high-quality results at a fraction of the time of RePaint\cite{repaint}. }
    \label{fig:res2}
\end{figure}

\begin{figure}[htbp]
  \centering
    \includegraphics[width=0.9\linewidth]{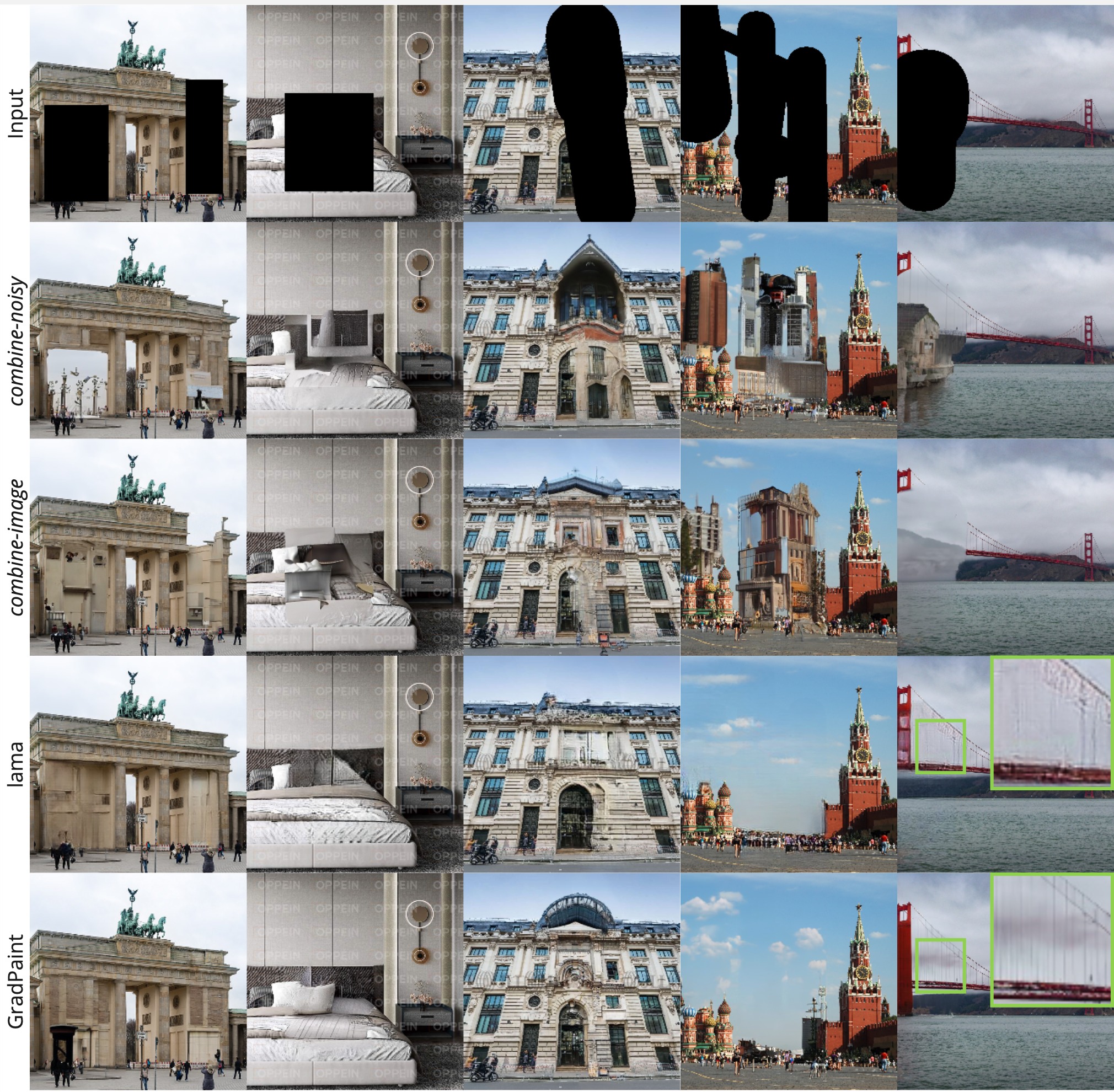}
    \caption{In-the-wild images for models trained on Places2. Note that \emph{combine-image}, \emph{combine-noisy} and GradPaint all use the same noise map for initialization. Note that LaMa was specifically trained using similar masks, contrary to our method.}
    \label{fig:places}
\end{figure}

\begin{figure}[ht]
  \centering
    \includegraphics[width=0.85\linewidth]{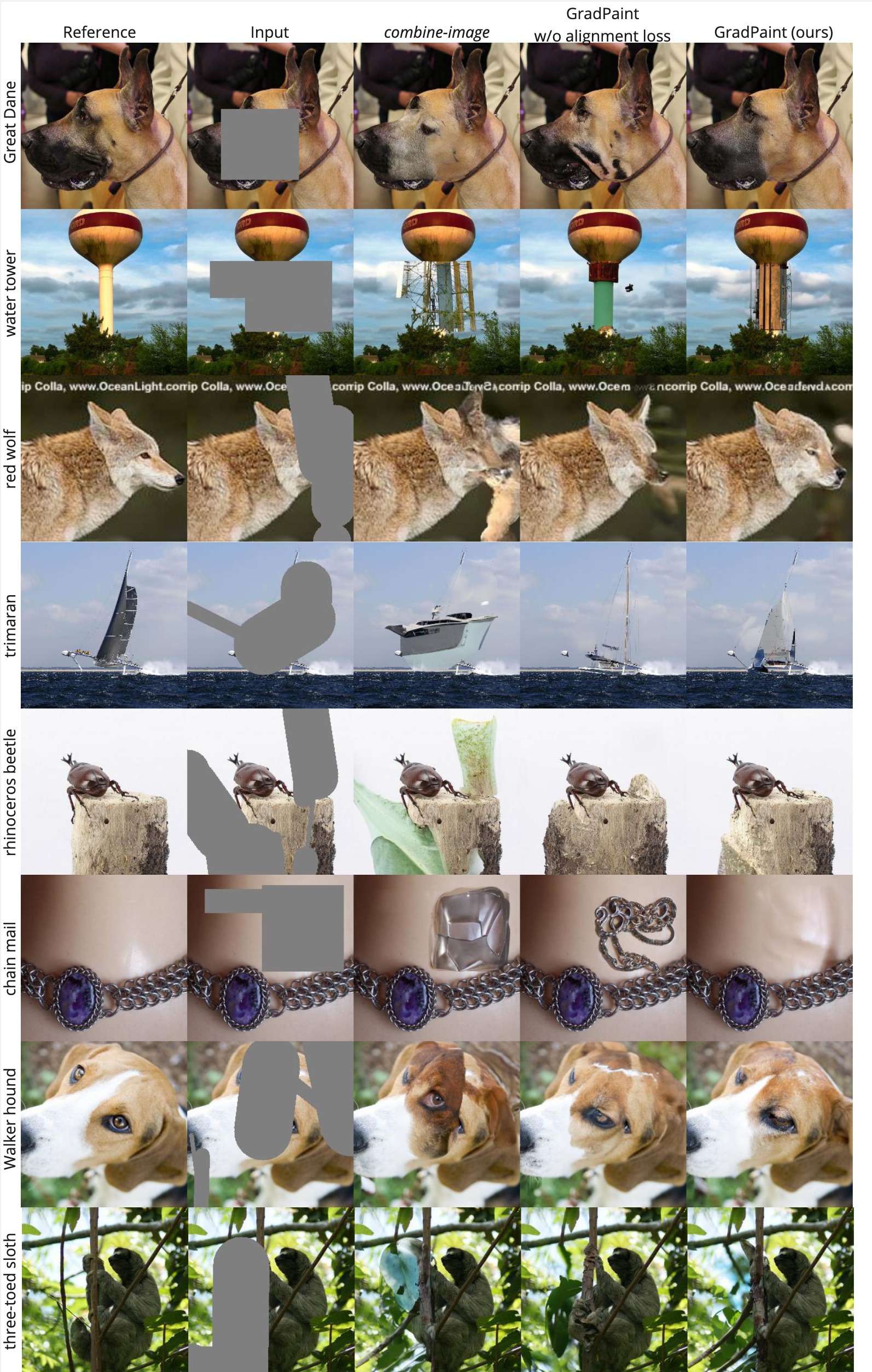}
    \caption{Qualitative results for select images of ImageNet dataset. Baseline \emph{combine-image} produces  images with visible artifacts. Our gradient update using only the masked MSE loss improves the ``copy-paste" effect, while the alignment loss produces better aligned transitions.}
    \label{fig:vis_ablation}
\end{figure}

\section{Qualitative Evaluation}

\noindent \paragraph{Image-space Diffusion Models}

Figs.~\ref{fig:res2} and~\ref{fig:places} and  show qualitative results using our method for Places2 and ImageNet pre-trained models, 
respectfully. Note that without the gradient-guidance of GradPaint, generations are unable to harmonize well.
Images produced by RePaint~\cite{lugmayr2022repaint} often lack global coherence, like the missing spider web in Fig.~\ref{fig:res2}. 
Our method produces globally and locally harmonized images, without the heavy computation cost of~\cite{lugmayr2022repaint} nor the 
specific supervised training of \cite{lama}. Note that we selected images where the \textit{thick} masks masked out key parts 
of the input image to better appreciate the different results.

Fig.~\ref{fig:vis_ablation} shows qualitative results on ImageNet-trained 
 guided diffusion model for different components of our method. We note that the baseline \emph{combine-image} is
  biased by the class-conditioning of the model without taking into account the context, like for the ``red wolf" class. 
  Adding gradient update as well as the alignment loss produces generations harmonized with the surrounding context.

\noindent \paragraph{Latent Diffusion Models}

\def\myim#1{\includegraphics[width=0.7\linewidth]{images/#1}}
\begin{figure}[h]
    \centering
    \renewcommand{\arraystretch}{1} 
    \setlength{\tabcolsep}{10pt} % tried 38; tried 30 ; tried 25
  \begin{tabular}{ccccc}
  \scriptsize
  Prompt  &  \hspace{0.2cm} \scriptsize Original Image & \scriptsize
  Mask & \scriptsize Baseline & \scriptsize GradPaint \\
  \end{tabular}
  \setlength{\tabcolsep}{5pt}
  \begin{tabular}{p{1.5cm}c} 
  %\quad Prompt  & Original Image \qquad \quad \quad Mask \qquad \quad \quad Baseline \qquad \quad GradPaint \\
  \vspace{-1.5cm}
  \begin{minipage}[c]{1.4cm}
    \scriptsize
  a group of ducks are sitting on a lake
  \end{minipage} & \myim{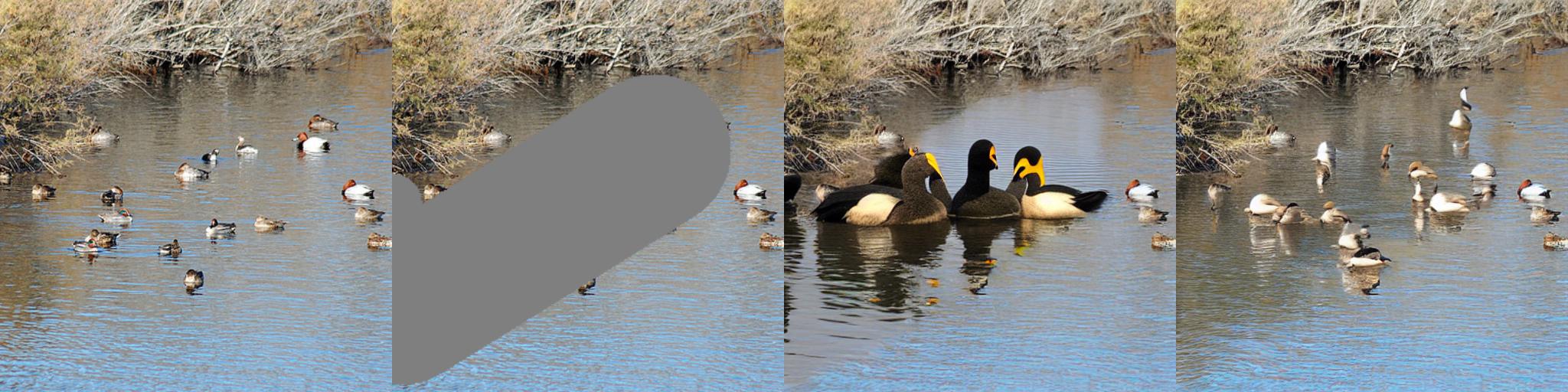} \\
  \vspace{-1.5cm}
  \begin{minipage}[c]{1.4cm}
    \scriptsize
  A person riding a motorcycle on a roadway.
  \end{minipage} & \myim{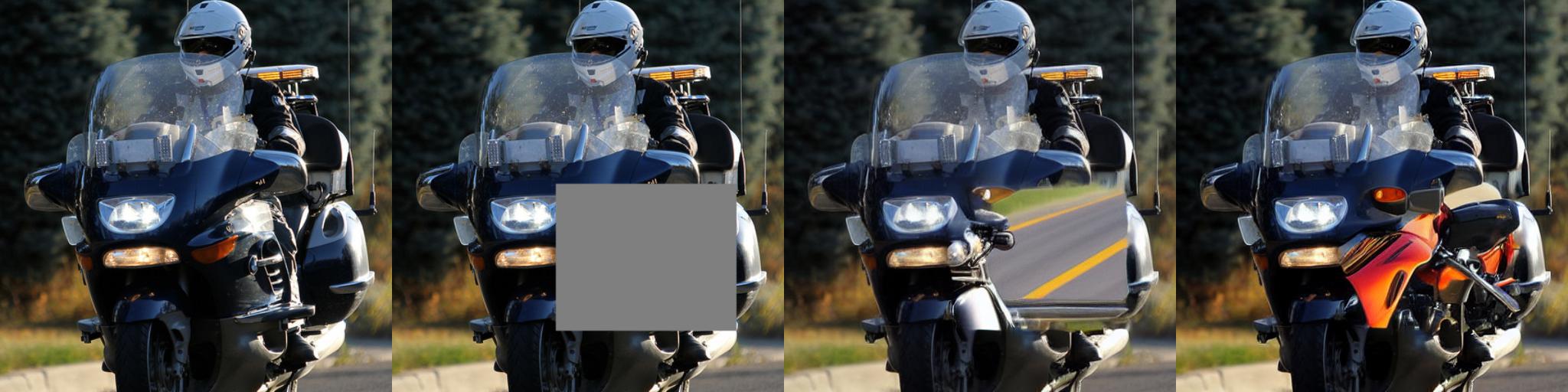} \\
  \vspace{-1.5cm}
  \begin{minipage}[c]{1.4cm}
    \scriptsize
  sheep look towards the camera as the stand in a field
  \end{minipage} & \myim{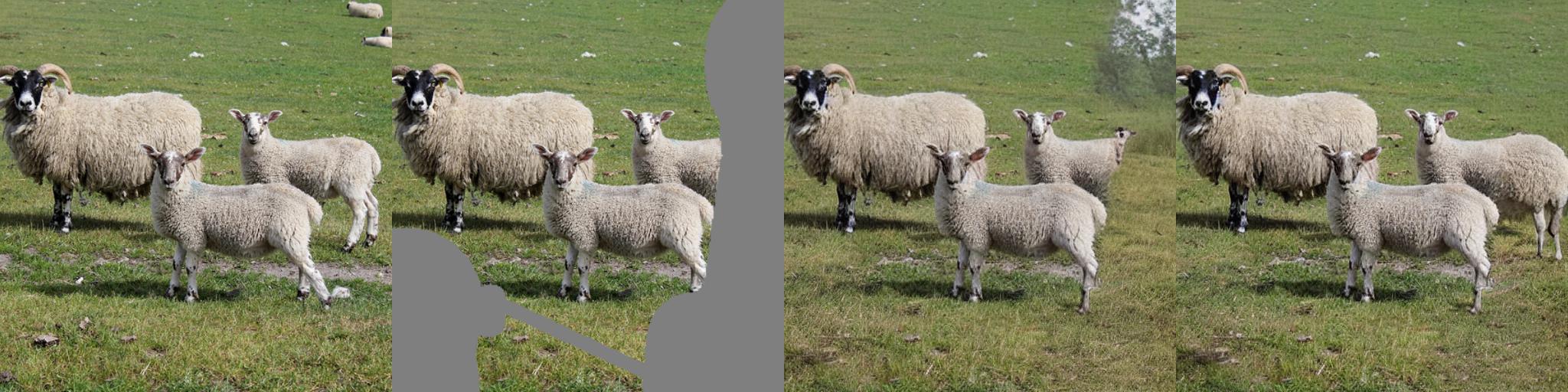} \\
  \vspace{-1.5cm}
  \begin{minipage}[c]{1.4cm}
    \scriptsize
  a bench leaves snow trees and some branches 
  \end{minipage} & \myim{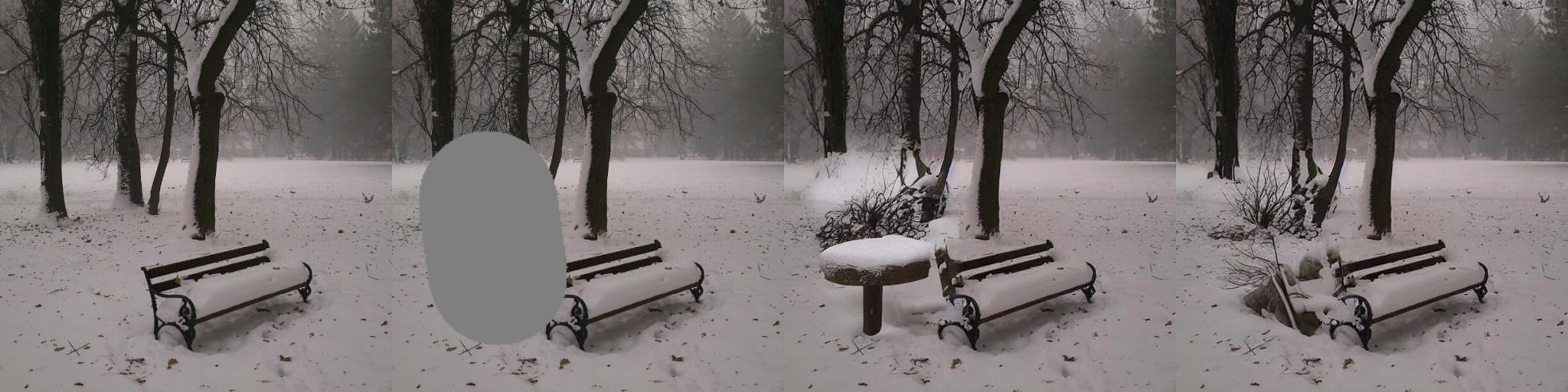} \\
  \end{tabular}
\caption{Qualitative results using Stable Diffusion on COCO. As we can see, our method successfully corrects unharmonized inpainted images.}
\label{fig:stable_diffusion_qual}
\end{figure}

Fig.~\ref{fig:stable_diffusion_qual} shows visual examples of our method using Stable Diffusion on COCO. 
Our method produces realistic and harmonized results compared to the baseline method. We  provide further results on ImageNet in the Appendix~\ref{sec:appendix_qualitative}.

\section{Impact of mask distribution} \label{ood_mask}

\begin{figure}[htbp]
  \centering
    \includegraphics[width=\linewidth]{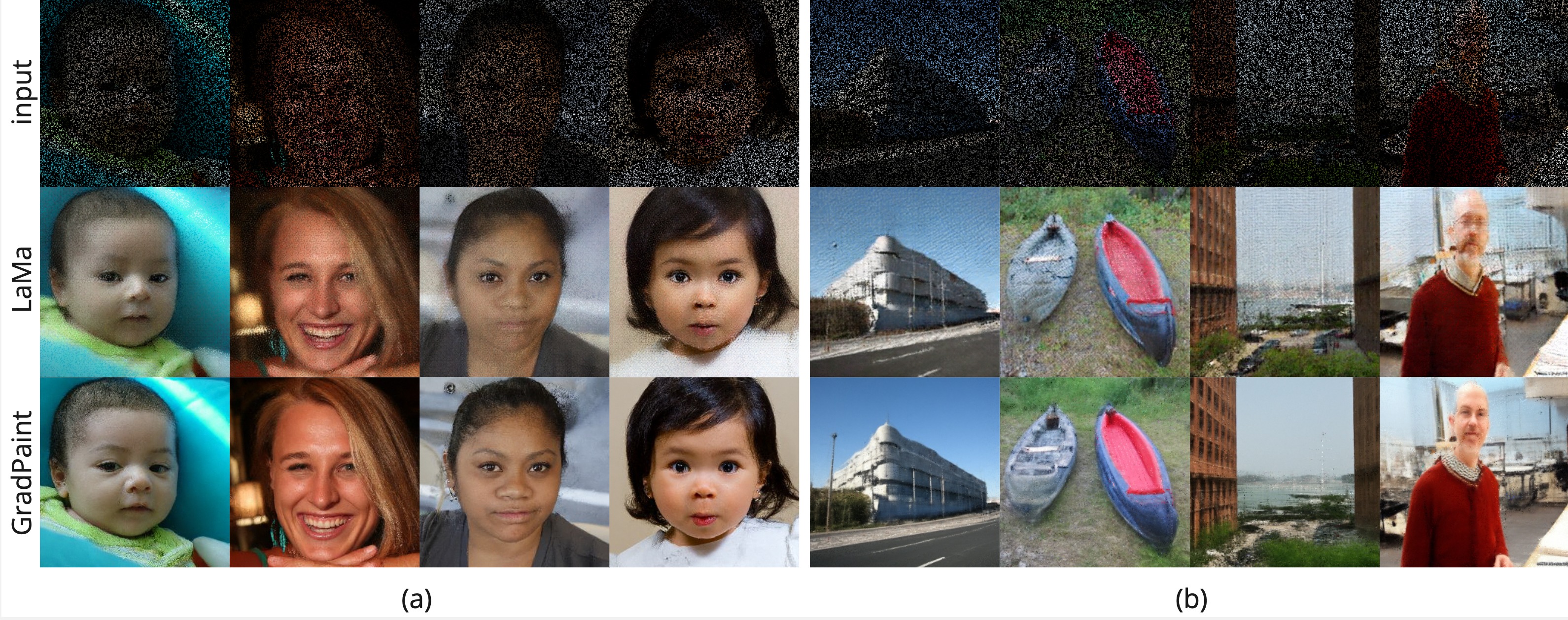}
    \caption{Uncurated results of our method compared to LaMa on out-of-distribution masks. We show images from (a) FFHQ and (b) Places2. LaMa fares poorly on masks outside of the training distribution. Best viewed zoomed and in color.}
    \label{fig:ood}
\end{figure}

Training-free is particularly advantageous as such a method is is agnostic to any 
pre-defined mask distribution to train on, contrary to training-based methods. We 
illustrate this by comparing our method to ~\cite{lama} on masks outside of 
their pre-defined training distribution. Specifically, we create masks where each 
pixel has a 80\% chance of being masked, masking considerable portions of the image. 
As we can see in Fig.~\ref{fig:ood}, \cite{lama} produces low-quality
 results while our method produces realistic images. This is confirmed quantitatively in Tab.~\ref{abnormal_tab}.

\begin{table}[H]
\centering
\begin{tabular}{|l|ll|ll|}
\hline
                   & \multicolumn{2}{c|}{CelebaHQ} & \multicolumn{2}{c|}{Places2} \\ \hline
                   & \multicolumn{1}{l|}{FID$\downarrow$}             & LPIPS$\downarrow$          & \multicolumn{1}{l|}{FID$\downarrow$}            & LPIPS$\downarrow$          \\ \hline
\rowcolor[gray]{0.7}
COPY (oracle)               & \multicolumn{1}{l|}{4.29}           & 0.0          & \multicolumn{1}{l|}{6.47}          & 0.0          \\ \hline

GREYFILL               & \multicolumn{1}{l|}{403.23}           & 1.06          & \multicolumn{1}{l|}{282.01}          & 1.09          \\ \Xhline{4\arrayrulewidth}

LaMa               & \multicolumn{1}{l|}{74.47}           & 0.517          & \multicolumn{1}{l|}{52.63}          & 0.320          \\ \hline

GradPaint & \multicolumn{1}{l|}{\textbf{44.87}}  & \textbf{0.170} & \multicolumn{1}{l|}{\textbf{27.17}} & \textbf{0.277} \\ \hline

\end{tabular}
\caption{Quantitative results comparing our train-free method to a training-based method (LaMa) on CelebaHQ and Places2 using masks outside of LaMa's training distribution.}
\label{abnormal_tab}
\vspace{1.5cm}
\end{table}

\section{Conclusion}

We have presented GradPaint, a training-free algorithm that guides the generative process of diffusion to better perform inpainting operations when given real images. GradPaint improves upon baselines by better harmonizing generated content inside the inpainting mask with known regions of the input image, which is done via gradient descent computed from a dedicated harmonization loss. Extensive qualitative and quantitative experiments demonstrate the superiority of our method, which is able to outperform methods trained specifically for inpainting.

It is important to note that many open-source diffusion models are trained with large amounts of web-scraped data, thus inheriting their biases. Applying our method onto these models could potentially reinforce harmful cultural biases. We believe open-sourcing editing algorithms in a research context contributes to a better understanding of these biases and will aid the community to mitigate them in the future.

\clearpage
{\small
\bibliographystyle{ieee_fullname}
\bibliography{asyabib}
}
\appendix

\onecolumn

\section{Details on Assets}
\label{sec:GradPaint assets}

Table \ref{tab:linkschap3} (links) and Table \ref{tab:licenceschap3} (licences)
list the assets we used in this work.

\begin{table*}[h]
\centering
\small
\begin{tabular}{lll}
\toprule
\textbf{Asset Name} & \textbf{Link} \\
\midrule
% datasets 
CelebA & https://mmlab.ie.cuhk.edu.hk/projects/CelebA.html \\
FFHQ &  https://github.com/NVlabs/ffhq-dataset \\
Places2 &  http://places2.csail.mit.edu/download.html \\
ImageNet & https://www.image-net.org \\
COCO & https://cocodataset.org/ \\
Guided Diffusion & https://github.com/openai/guided-diffusion \\
Latent Diffusion & https://github.com/CompVis/latent-diffusion \\
Stable Diffusion & https://github.com/CompVis/stable-diffusion \\
LaMa & https://github.com/advimman/lama \\
Palette & https://github.com/Janspiry/Palette-Image-to-Image-Diffusion-Models \\
LPIPS &  https://github.com/richzhang/PerceptualSimilarity\\
FID & https://github.com/mseitzer/pytorch-fid \\ % code and moels
FFHQ pre-trained model & https://github.com/yandex-research/ddpm-segmentation \\ % just model
RePaint & https://github.com/andreas128/RePaint\\
MCG & https://github.com/HJ-harry/MCG\_diffusion\\
\bottomrule
\end{tabular}
\caption{List of asset links.}
\label{tab:linkschap3}
\end{table*}

\begin{table*}[h]
\centering
\small
\begin{tabular}{lll}
\toprule
\textbf{Asset Name} & \textbf{Asset type} & \textbf{License} \\
\midrule
CelebA & Images & CC BY-NC-SA 4.0 License \\
FFHQ & Images &  https://github.com/NVlabs/ffhq-dataset/blob/master/LICENSE.txt \\
Places2 &  Images & Creative Commons Attribution 4.0 International \\
ImageNet & Images & https://www.image-net.org/download.php \\
COCO & Images & Creative Commons Attribution 4.0 License \\
Guided Diffusion & Code and Models & MIT License\\
Latent Diffusion & Code and Models & MIT License \\
Stable Diffusion & Code and Model & CreativeML Open RAIL-M\\
LaMa & Code and Models & Apache License 2.0 \\
Palette & Code and Models & MIT License \\
LPIPS & Code and Models & BSD-2-Clause License \\
FID & Code and Models & Apache-2.0 License \\
FFHQ pre-trained model & Model & MIT License \\% just model
RePaint & Code & CC BY-NC-SA 4.0 License\\
MCG & Code & Apache License 2.0\\
\bottomrule
\end{tabular}
\caption{List of asset licenses.}
\label{tab:licenceschap3}
\end{table*}

\section{Inference Time Study}\label{speed_section}

\begin{figure*}[h]
    \centering
    \begin{subfigure}[t]{0.5\linewidth}
        \centering
        \includegraphics[height=2.3in]{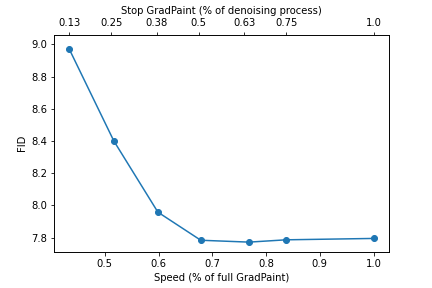}
        \caption{FID vs. Time tradeoff}
    \end{subfigure}%
    ~ 
    \begin{subfigure}[t]{0.5\linewidth}
        \centering
        \includegraphics[height=2.3in]{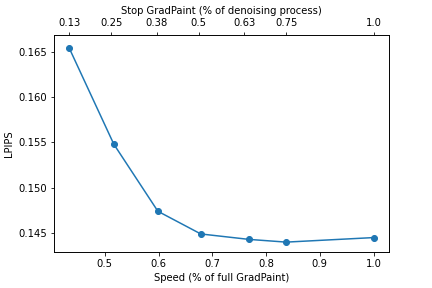}
        \caption{LPIPS vs. Time tradeoff}
    \end{subfigure}
    \caption{Performance vs. Time tradeoff, performed on ImageNet. We performed various settings where we early stopped the gradient calculation of GradPaint at $13\%, 25\%, 38\%, 50\%, 63\%, 75\%$ of the denoising process. The most important gains of GradPaint occurs thanks to the early gradient-guidance, and after $63\%$, gradient guiding no longer helps performance. A reasonable early-stopping at $50\%$ of the denoising process reduces the initial time of GradPaint by $33\%$.}
    \label{timsvsacc}
\end{figure*}

Our method requires approximately 3x the compute time of the  gradient-free sampling baselines \emph{combine-image} and 
\emph{combine-noisy}, explained by the added backpropagation of every step of the denoising process. As illustrated in Fig.~\ref{fig:intuition}, 
our method succeeds because the prediction $\hat{x}_0$ is aligned and harmonized with the input image early
 on in the diffusion process. We can improve the time of GradPaint by early-stopping the gradient calculation and letting the 
 rest of the denoising process run as in \emph{combine-image}. If the harmonization succeeded well-enough early on, then subsequent
  gradient calculations may not be necessary later in the denoising process, saving valuable compute time. We performed these 
  experiments for various early stopping times of the total denoising process. Fig.~\ref{timsvsacc} shows the performance in terms
   of LPIPS and FID for the various settings. Performing GradPaint after $63\%$ of the denoising process doesn't improve performance, 
   and significantly increases compute time. Stopping the gradient calculation at about $50\%$ of the denoising process achieves the 
   most important gains in performance and allows us to reduce the initial GradPaint time by $33\%$. This early stopping is referred 
   to as "GradPaint-Fast" below.

Lastly, we compare the inference times of various methods in Table~\ref{tab:comp_times}.
Our method is faster than other diffusion-based methods, even those specifically trained for inpainting (Palette), 
especially when applied to LDMs. We remark that RePaint~\cite{lugmayr2022repaint} takes $313s$ (over 5 minutes!) per image, 
making it unfitting for practical use.

\begin{table}[h]
    \centering
    \begin{tabular}{|l|c|c|c|}
    \hline
    Method           & \multicolumn{1}{l|}{Training-Free?} & Diffusion-based? & Inference time per image (s)  \\ \hline
    LaMa             & \multicolumn{1}{l|}{} &               & 0.02                                                                          \\ \hline
    Palette          & \multicolumn{1}{l|}{} & \checkmark              & 56                                                                            \\ \hline
    RePaint          & \checkmark  & \checkmark                                 & 313                                                                           \\ \hline
    MCG              & \checkmark   & \checkmark                                & 52                                                                            \\ \hline
    GradPaint        & \checkmark  & \checkmark                                 & 66                                                                            \\ \hline
    GradPaint-Fast   & \checkmark  & \checkmark                                 & 44                                                                          \\ \hline
    GradPaint-Latent & \checkmark & \checkmark                                  & 7.2                                                                           \\ \hline
    \end{tabular}
\caption{Inference times for various methods. As we can see, our method is faster than other diffusion-based methods, especially when applied to Latent Diffusion Models}
\label{tab:comp_times}
\end{table}\textbf{}

\section{Inpainting with diversity}

\begin{figure}[htbp]
  \centering
    \includegraphics[width=\linewidth]{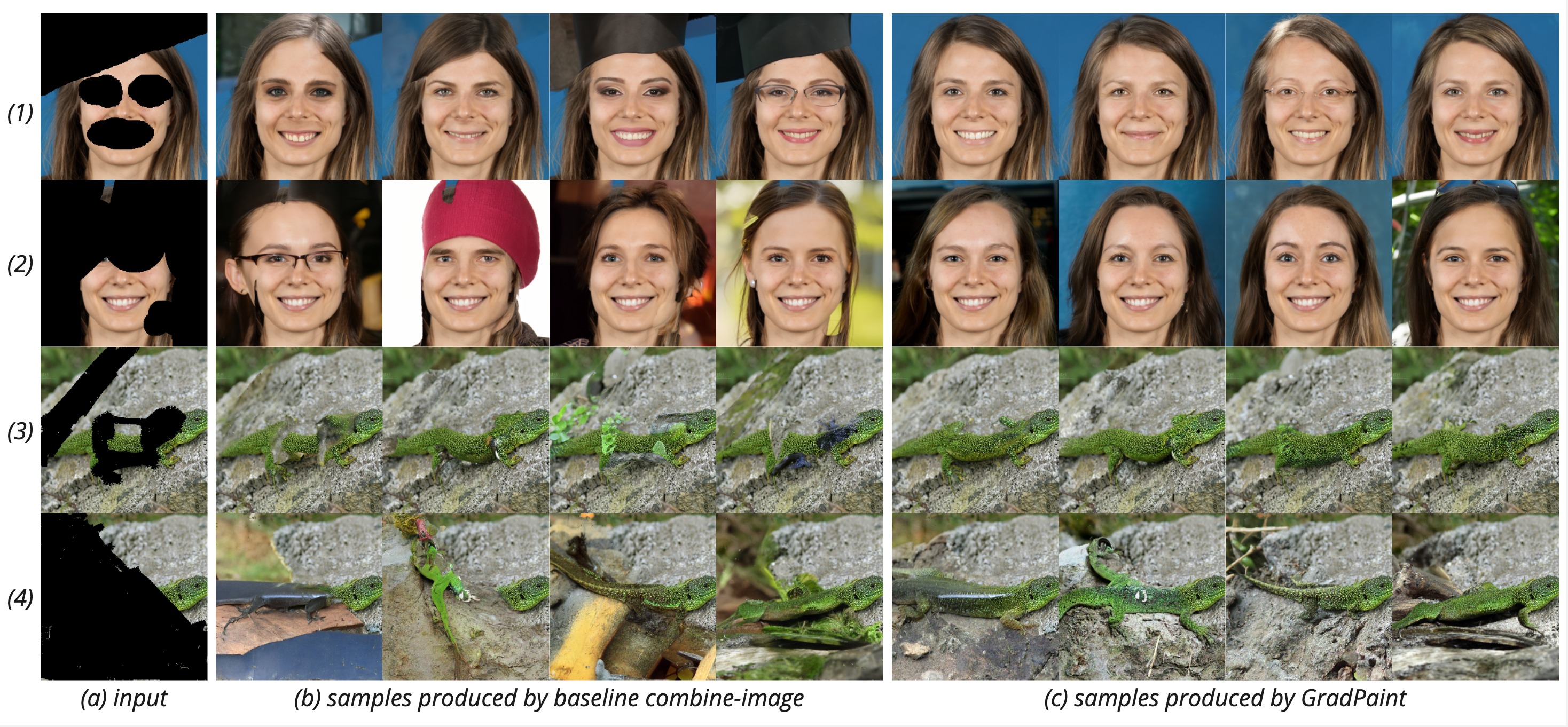}
    \caption{Diversity of select samples from 30 random samples. Images which are well-harmonized are less diverse, e.g. in (1), our method is encouraged to produce a blue background using the small available context, disregarded by the baseline method which produces more diverse but unharmonized backgrounds. Matching columns of samples of (b) and (c) are initialized identically.}
    \label{fig:diversity}
\end{figure}

A method which well-harmonizes an image for the inpainting task will, naturally, produce less diverse images. This can be illustrated in Fig.~\ref{fig:diversity}. Because the baseline method poorly utilizes the surrounding context, samples are diverse but not realistic. On the other hand, our method leverages the surrounding context which produces a more natural result, implying more consistent generations. 
To further analyze the diversity of our samples, we used one single input image and generated 500 outputs, comparing our method with the baseline \emph{combine-image}. We then extracted image features using the inception v3 pre-trained model~\cite{inceptionv3} and display the average of the variance of these features. Tab.~\ref{tab:diversitytab} shows the results. Our method gives diverse samples as the input mask sizes increases, but limits absurd and unharmonized results.

\begin{table}[]
\centering
\begin{tabular}{|l|l|l|l|l|}
\hline
\multicolumn{1}{|c|}{ mask coverage (\% of im.)} & \multicolumn{1}{c|}{10\%} & \multicolumn{1}{c|}{25\%} & \multicolumn{1}{c|}{50\%} & \multicolumn{1}{c|}{75\%} \\ \hline
\textit{combine-image}                            & 2.3                       & 5.5                       & 10.5                      & 11.9                      \\ \hline
GradPaint                                         & 1.9                       & 4.4                       & 8.7                       & 9.6                       \\ \hline
\end{tabular}
\caption{Variance of image features (in $10^{-3}$) for GradPaint compared to the baseline method, for 500 generated results using the same input image and increasing-sized masks.  Our method expectedly produces more diverse images as the inpainting mask increases, nearing the diversity of the baseline for large masks.}
\label{tab:diversitytab}
\vspace{-.45cm}
\end{table}

\section{Additional Quantitative and Qualitative Results}

Tab.~\ref{thin_med_results} compares various training-based (LaMa and Palette) and training-free (Repaint, \emph{combine-noisy}, \emph{combine-image} MCG and GradPaint (ours)) methods for \emph{thin} and \emph{medium} masks which LaMa used for training. As we can see, our method generally achieves the best FID score and a comparable LPIPS score even with LaMa's fully supervised method.  

\begin{table}[h]
\small
\centering
\begin{tabular}{|l|cccccc|cccccc|}
\hline
\multicolumn{1}{|r|}{Dataset} &
  \multicolumn{4}{c|}{FFHQ (pretrained on CelebaHQ)} &
  \multicolumn{4}{c|}{CelebaHQ (pretrained on FFHQ)} \\ \hline
\multicolumn{1}{|r|}{Masks} &
  \multicolumn{2}{c|}{Thin} &
  \multicolumn{2}{c|}{Medium} &
  \multicolumn{2}{c|}{Thin} &
  \multicolumn{2}{c|}{Medium} \\ \hline
\multicolumn{1}{|r|}{Metrics} &
  \multicolumn{1}{c|}{FID$\downarrow$} &
  \multicolumn{1}{c|}{LPIPS$\downarrow$} &
  \multicolumn{1}{c|}{FID$\downarrow$} &
  \multicolumn{1}{c|}{LPIPS$\downarrow$} &
  \multicolumn{1}{c|}{FID$\downarrow$} &
  \multicolumn{1}{c|}{LPIPS$\downarrow$} &
  \multicolumn{1}{c|}{FID$\downarrow$} &
  \multicolumn{1}{c|}{LPIPS$\downarrow$} \\ \hline %\hhline{|=|=|=|=|=|=|=|=|=|=|=|=|=|}
  \rowcolor[gray]{0.7}
COPY (oracle) &
  \multicolumn{1}{c|}{4.29} &
  \multicolumn{1}{c|}{0} &
  \multicolumn{1}{c|}{4.29} &
  \multicolumn{1}{c|}{0} &
  \multicolumn{1}{c|}{3.01} &
  \multicolumn{1}{c|}{0} &
  \multicolumn{1}{c|}{3.01} &
  \multicolumn{1}{c|}{0} 
  \\ \hline
GREYFILL &
  \multicolumn{1}{c|}{154.6} &
  \multicolumn{1}{c|}{0.353} &
  \multicolumn{1}{c|}{98.75} &
  \multicolumn{1}{c|}{0.250} &
  \multicolumn{1}{c|}{173.65} &
  \multicolumn{1}{c|}{0.3770} &
  \multicolumn{1}{c|}{124.32} &
  \multicolumn{1}{c|}{0.2631} 
  \\ \Xhline{4\arrayrulewidth}
LaMa&
  \multicolumn{1}{c|}{\textbf{6.22}} &
  \multicolumn{1}{c|}{\textbf{0.041}} &
  \multicolumn{1}{c|}{{\underline{5.61}}} &
  \multicolumn{1}{c|}{\textbf{0.052}} &
  \multicolumn{1}{c|}{n/a} &
  \multicolumn{1}{c|}{n/a} &
  \multicolumn{1}{c|}{n/a} &
  \multicolumn{1}{c|}{n/a} 
  \\ \hline
Palette &
  \multicolumn{1}{c|}{6.78} &
  \multicolumn{1}{c|}{{\underline{0.049}}} &
  \multicolumn{1}{c|}{6.78} &
  \multicolumn{1}{c|}{{\underline{0.068}}} &
  \multicolumn{1}{c|}{n/a} &
  \multicolumn{1}{c|}{n/a} &
  \multicolumn{1}{c|}{n/a} &
  \multicolumn{1}{c|}{n/a}  
  \\ \Xhline{4\arrayrulewidth}
Repaint &
  \multicolumn{1}{c|}{13.25} &
  \multicolumn{1}{c|}{0.060} &
  \multicolumn{1}{c|}{12.21} &
  \multicolumn{1}{c|}{0.080} &
  \multicolumn{1}{c|}{8.23} &
  \multicolumn{1}{c|}{\textbf{0.047}} &
  \multicolumn{1}{c|}{8.14} &
  \multicolumn{1}{c|}{\underline{0.062}} 
  \\ \hline
MCG &
  \multicolumn{1}{c|}{7.71} &
  \multicolumn{1}{c|}{0.062} &
  \multicolumn{1}{c|}{5.97} &
  \multicolumn{1}{c|}{0.073} &
  \multicolumn{1}{c|}{8.43} &
  \multicolumn{1}{c|}{0.059} &
  \multicolumn{1}{c|}{6.52} &
  \multicolumn{1}{c|}{0.065} 
  \\ \hline
\begin{tabular}[c]{@{}l@{}}\textit{combine-noisy}\end{tabular} &
  \multicolumn{1}{c|}{13.48} &
  \multicolumn{1}{c|}{0.090} &
  \multicolumn{1}{c|}{9.35} &
  \multicolumn{1}{c|}{0.099} &
  \multicolumn{1}{c|}{ 11.1} &
  \multicolumn{1}{c|}{0.070} &
  \multicolumn{1}{c|}{9.78} &
  \multicolumn{1}{c|}{0.081} 
  \\ \hline
\begin{tabular}[c]{@{}l@{}}\textit{combine-image}\end{tabular} &
  \multicolumn{1}{c|}{11.19} &
  \multicolumn{1}{c|}{0.096} &
  \multicolumn{1}{c|}{7.49} &
  \multicolumn{1}{c|}{0.102} &
  \multicolumn{1}{c|}{\underline{7.89}} &
  \multicolumn{1}{c|}{0.080} &
  \multicolumn{1}{c|}{\underline{5.96}} &
  \multicolumn{1}{c|}{0.089} \\ \hline
GradPaint (ours) &
  \multicolumn{1}{c|}{{\underline{6.613}}} &
  \multicolumn{1}{c|}{0.060} &
  \multicolumn{1}{c|}{\textbf{5.39}} &
  \multicolumn{1}{c|}{{\underline{0.069}}} &
  \multicolumn{1}{c|}{\textbf{5.13}} &
  \multicolumn{1}{c|}{{\underline{0.051}}} &
  \multicolumn{1}{c|}{\textbf{4.29}} &
  \multicolumn{1}{c|}{\textbf{0.077}} \\ \hline
\end{tabular}
\vspace{1mm}
\caption{Evaluation of various methods on FFHQ and CelebaHQ datasets for \textit{thin} and \textit{medium} masks. Rows COPY (oracle) and GREYFILL are respectfully the lower and upper bounds for LPIPS and FID. LaMa and Palette are both training-based methods. RePaint, \emph{combine-noisy}, \emph{combine-image}, MCG and GradPaint are all training-free methods which all use the same model based on guided diffusion. Best score is shown \textbf{in bold} and second best \underline{underlined}.}
\label{thin_med_results}
\end{table}

\begin{figure*}
    \centering
    \includegraphics[width=\linewidth]{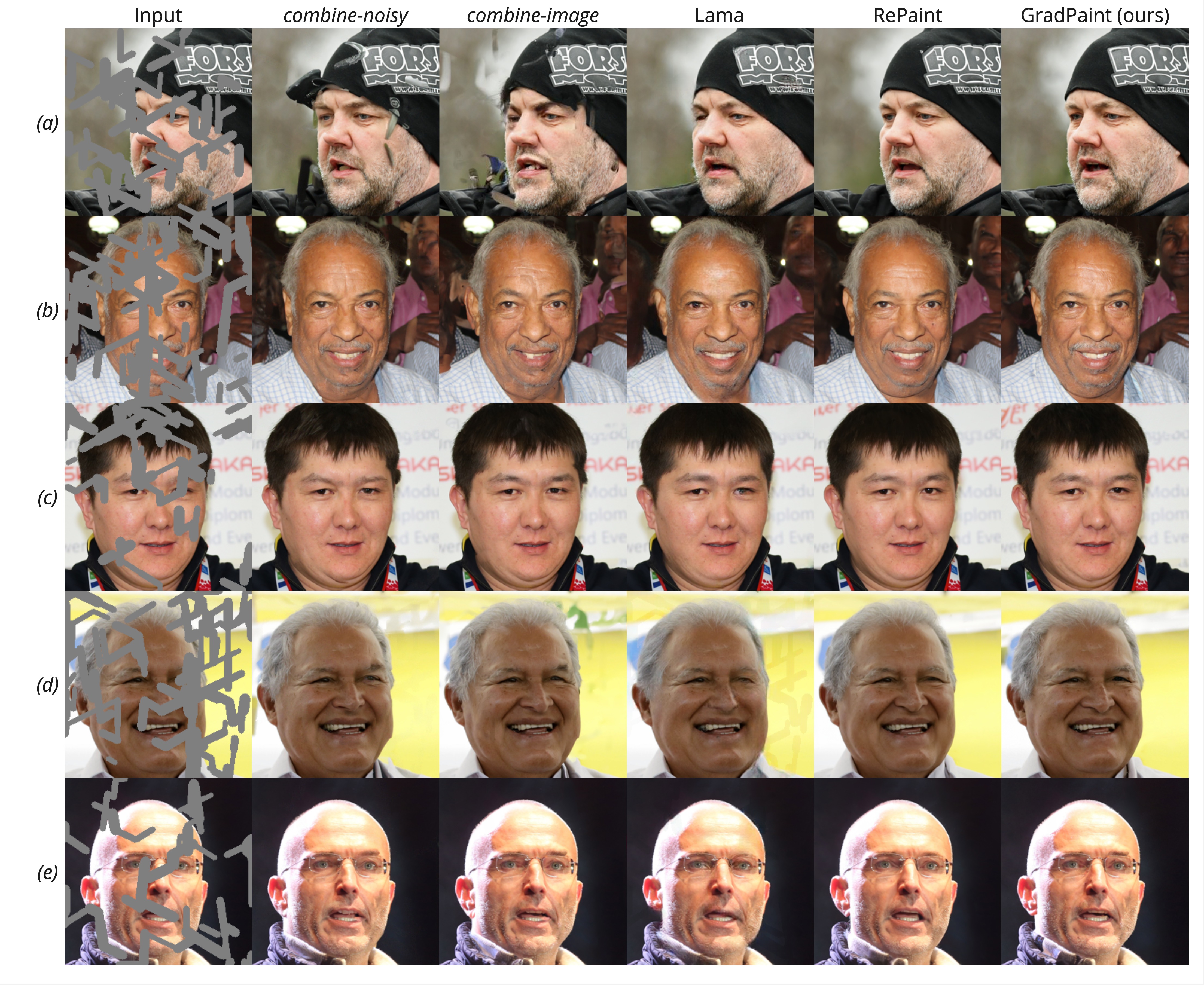}
    \caption{Qualitative results with thin masks. Best viewed zoomed and in color. In \emph{(a)}, notice the quality of the text of the hat, which our method most successfully reconstructs. Notice the background face on the right in \emph{(b)}, which only our method managed to decently generate. Finally, in \emph{(c)}, \emph{(d)}, and \emph{(e)}, methods \emph{combine-noisy} and \emph{combine-image} have harmonization issues (see eyes, cheek/eyebrow, and lips respectively), while our method is on par with the costly RePaint.}
    \label{fig:qualitative_thin}
\end{figure*}

\begin{figure*}
    \centering
    \includegraphics[width=\linewidth]{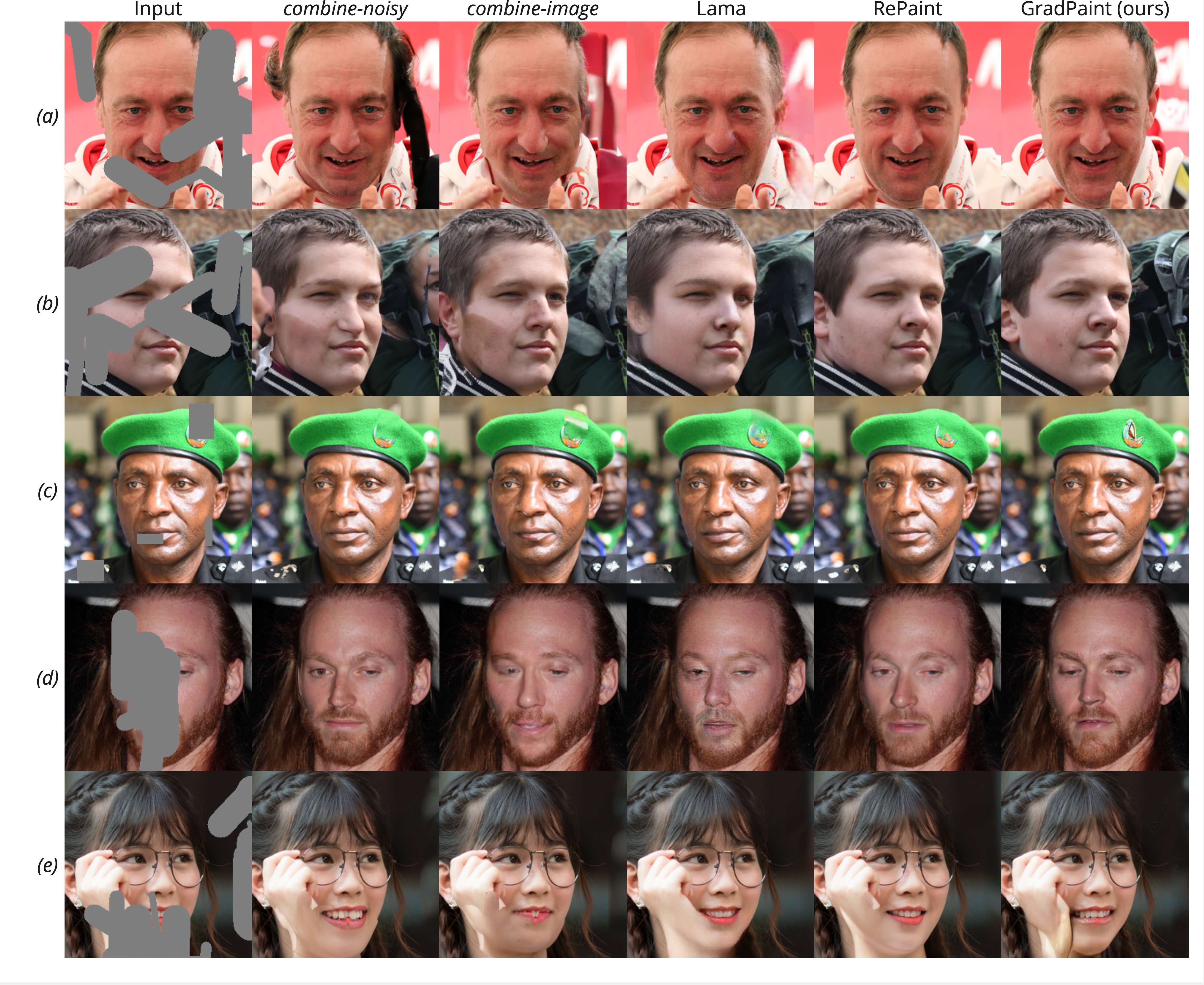}
    \caption{Qualitative results with medium masks. Best viewed zoomed and in color. Methods \emph{combine-noisy} and \emph{combine-image} have harmonization issues, particularly in \emph{(a)} and \emph{(b)}. Notice the badge on the hat in \emph{(c)} where most methods fail to realistically integrate it with the hat. In \emph{(d)}, our method most successfully generates an eye which matches with the input eye. Finally, all methods struggle with the difficult example \emph{(e)}, as it is poorly represented in the training distriubtion.  Nevertheless, our method is the sole one which attempts to reconstruct the hand as a separate entity from the face.}
    \label{fig:qualitative_thick}
\end{figure*}

Figs.~\ref{fig:qualitative_thin} and ~\ref{fig:qualitative_thick} show additional qualitative examples using \emph{thin} and \emph{medium} masks. With these smaller masks, the visual differences are sometimes harder to appreciate, and we recommend zooming when viewing results. We compare our method to the baselines \emph{combine-image}, \emph{combine-noisy}, LaMa, and RePaint. Baselines \emph{combine-noisy} and \emph{combine-image} clearly have harmonization issues. LaMa tends to have blurry artifacts typical of GANs in the generated areas. RePaint generally produces realistic generations, but which aren't always coherent with other elements of the image. Our method produces high-quality results at a much smaller computational cost than RePaint.

\begin{figure*}[htbp]
  \centering
    \includegraphics[width=0.95\linewidth]{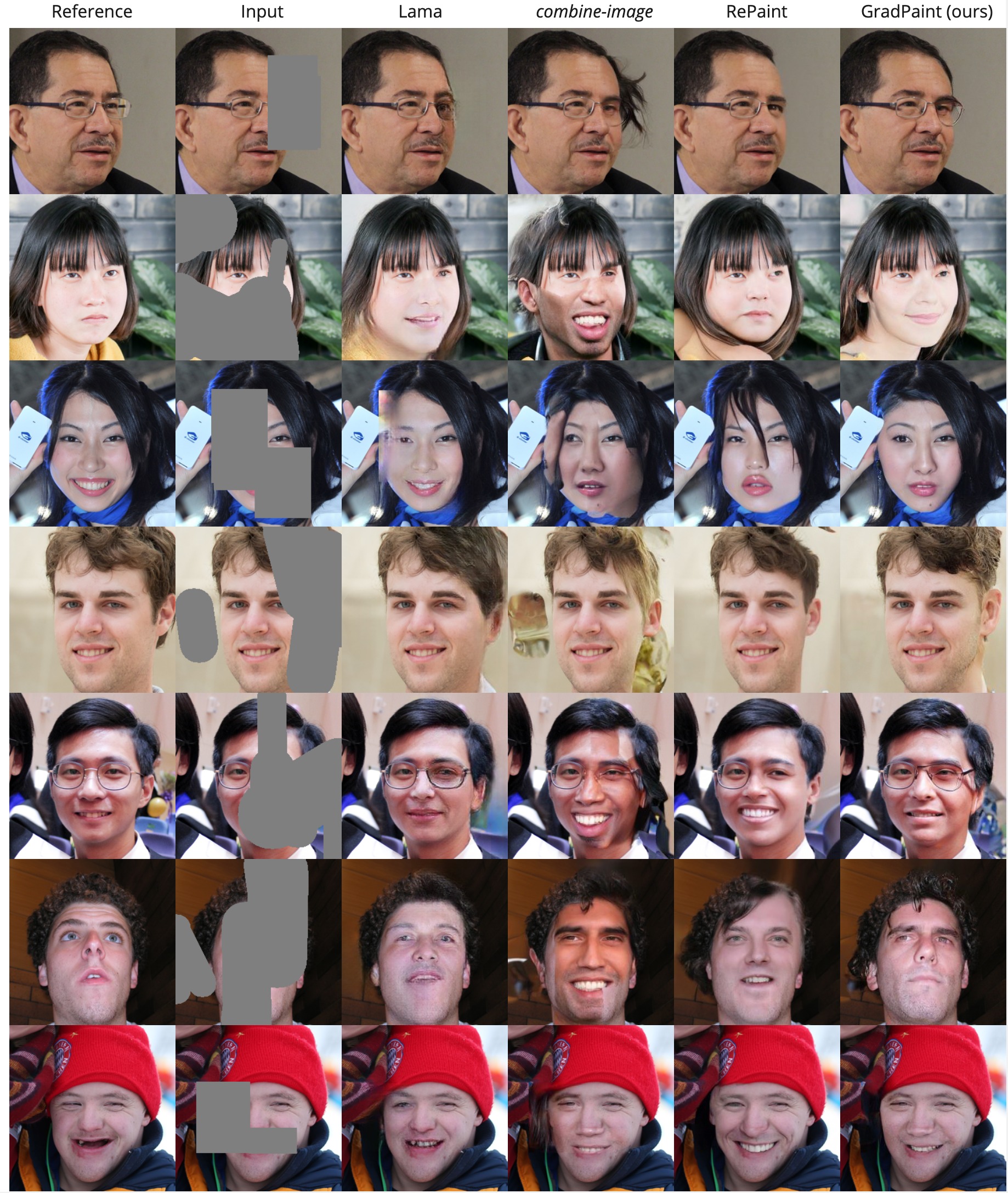}
    \caption{Inpainting results on select images from FFHQ. Best viewed zoomed and in color. Notice that while the overall reconstruction of LaMa is decent, zooming on the image unveils visible blurry and unrealistic artifacts typical of GANs. Results using the \emph{combine-image} baseline are of poor quality with an obvious ``copy/paste" effect at the mask. Notice that even with the 4500 forward passes necessary for RePaint, the overall image coherence often fails, despite the high-quality local generation.  Our method produces high-quality generation while matching with the global structure of the rest of the image. All models were trained on CelebAHQ dataset.}
    \label{fig:res_ffhq}
\end{figure*}

Fig.~\ref{fig:res_ffhq} shows additional select examples on FFHQ dataset. Notice that LaMa produces results close to the reference image, but generally containing blurry artifacts, explaining the generally higher LPIPS scores but lower FID scores than our method. RePaint's method, despite requiring over 5 minutes of compute time for a generation, produces smooth local changes but often fails to harmonize with the global structure of the image (rows 1, 2, 4, 5). Our method produces high-quality and harmonized results, especially visible when the generation requires surrounding context, such as rows 1 and 5 which necessitate generating glasses.

%will be displayed next page

\section{Additional Qualitative experiments with Latent Diffusion Models}\label{sec:appendix_qualitative}
\begin{figure*}
    \centering
    \includegraphics[width=0.625\linewidth]{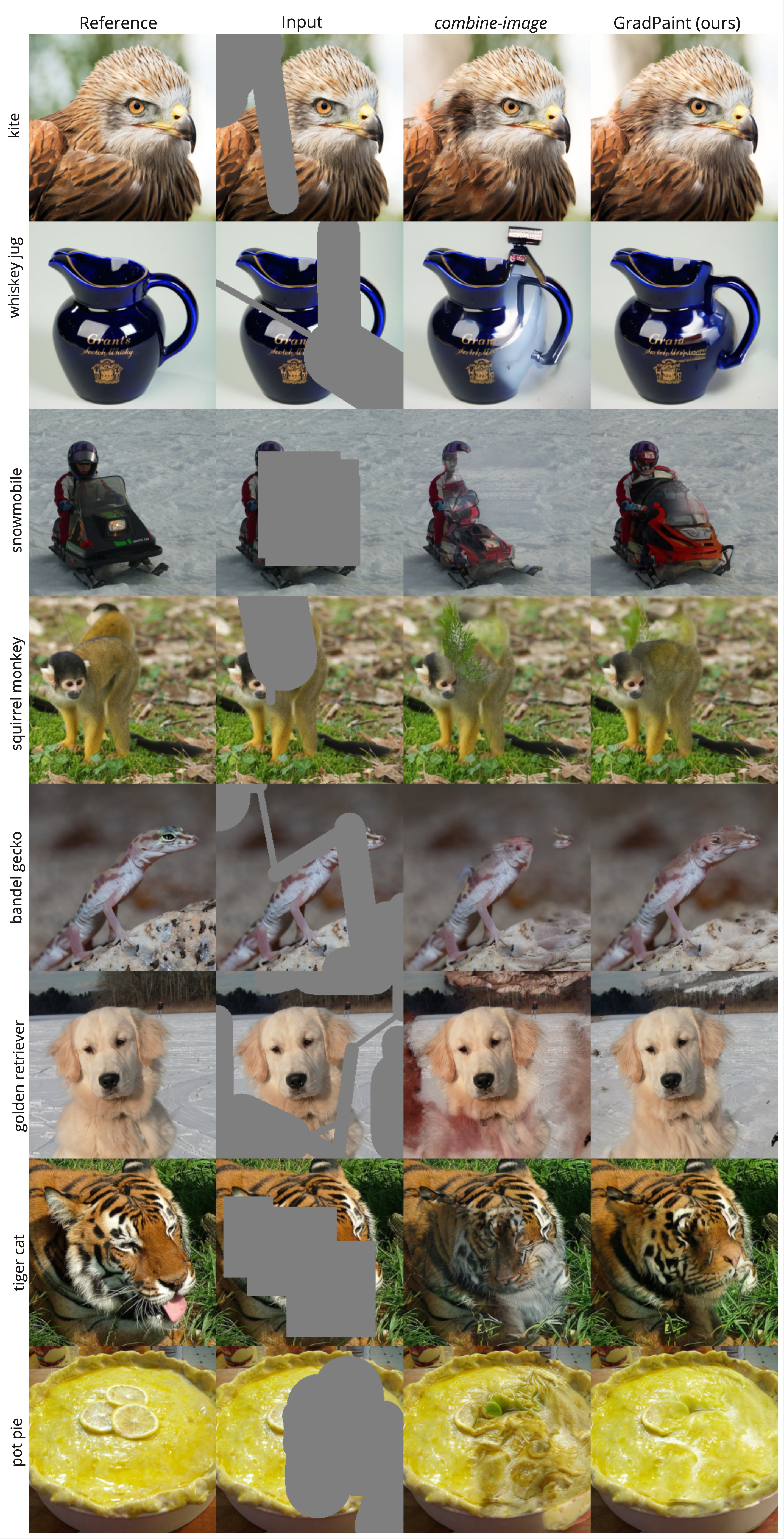}
    \caption{Qualitative examples from ImageNet with the latent diffusion model. Both the baseline and our algorithm are initialized with the same initial noise map.}
    \label{fig:qualitative_latent_imagenet}
\end{figure*}

Fig.~\ref{fig:qualitative_latent_imagenet} shows visual examples of our inpainting results using latent diffusion models. While the baseline method produces low-quality results, our method produces realistic and harmonized results.

\section{Uncurated Results and Limits}

\begin{figure*}[htbp]
  \centering
    \includegraphics[width=0.57\linewidth]{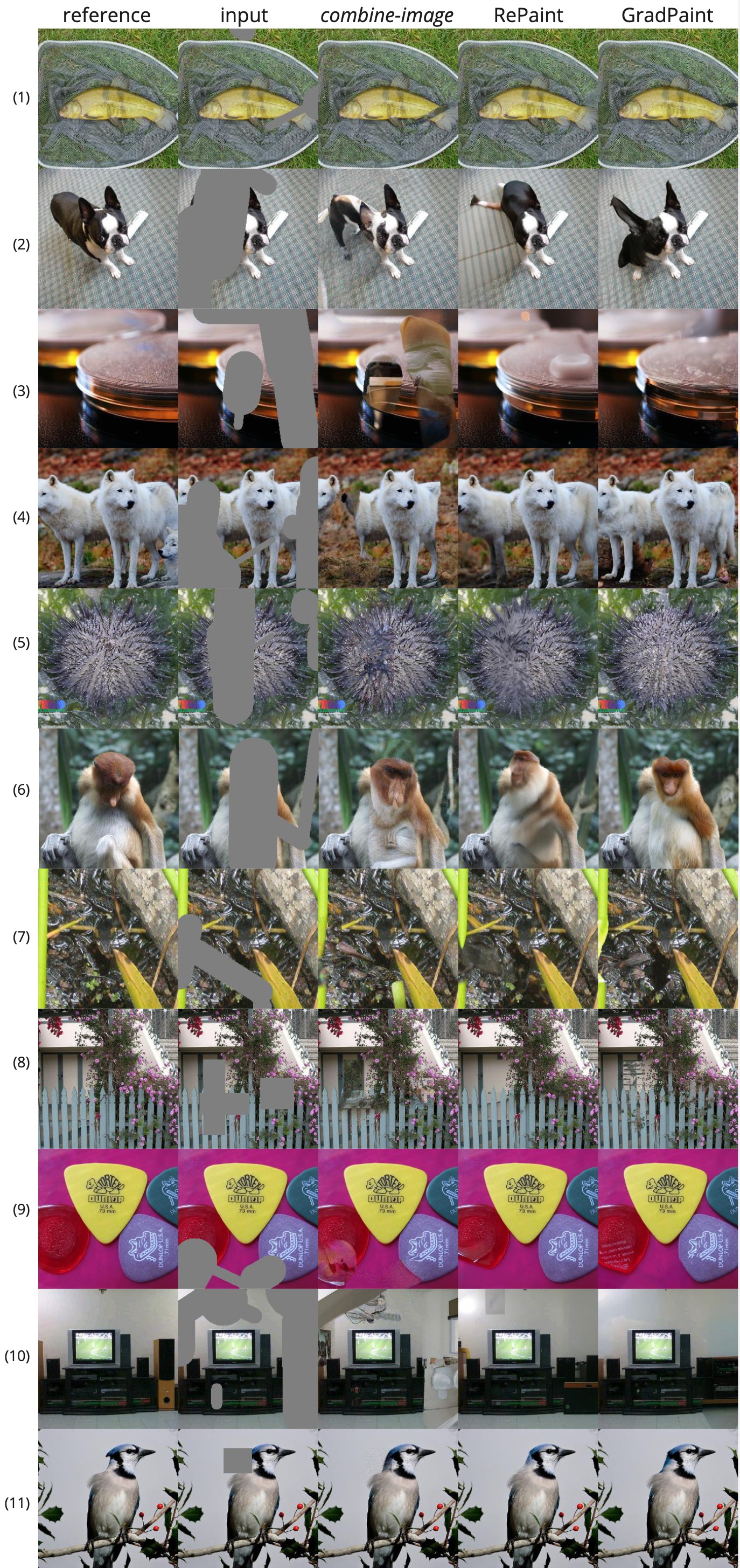}
    \caption{Uncurated results on ImageNet (1). Rows 1, 2, 4, 5, 6 display the superior ability of GradPaint to align fine details and global coherence.}
    \label{fig:uncur1}
\end{figure*}

\begin{figure*}[htbp]
  \centering
    \includegraphics[width=0.57\linewidth]{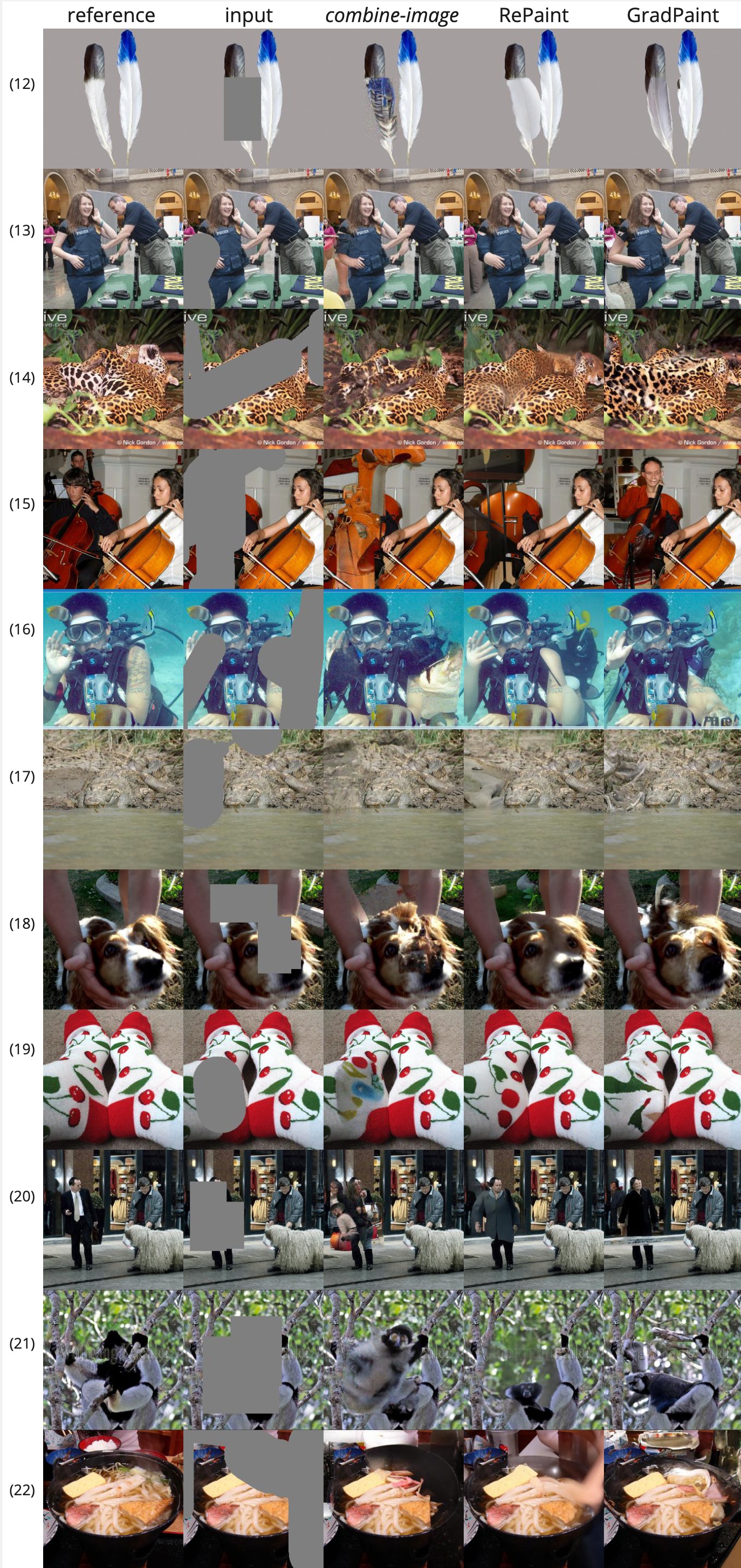}
    \caption{Uncurated results on ImageNet (2). GradPaint works well on alignement tasks (row 12), but sometimes struggles with non-standard input (row 14). }
    \label{fig:uncur2}
\end{figure*}

\begin{figure*}[htbp]
  \centering
    \includegraphics[width=0.57\linewidth]{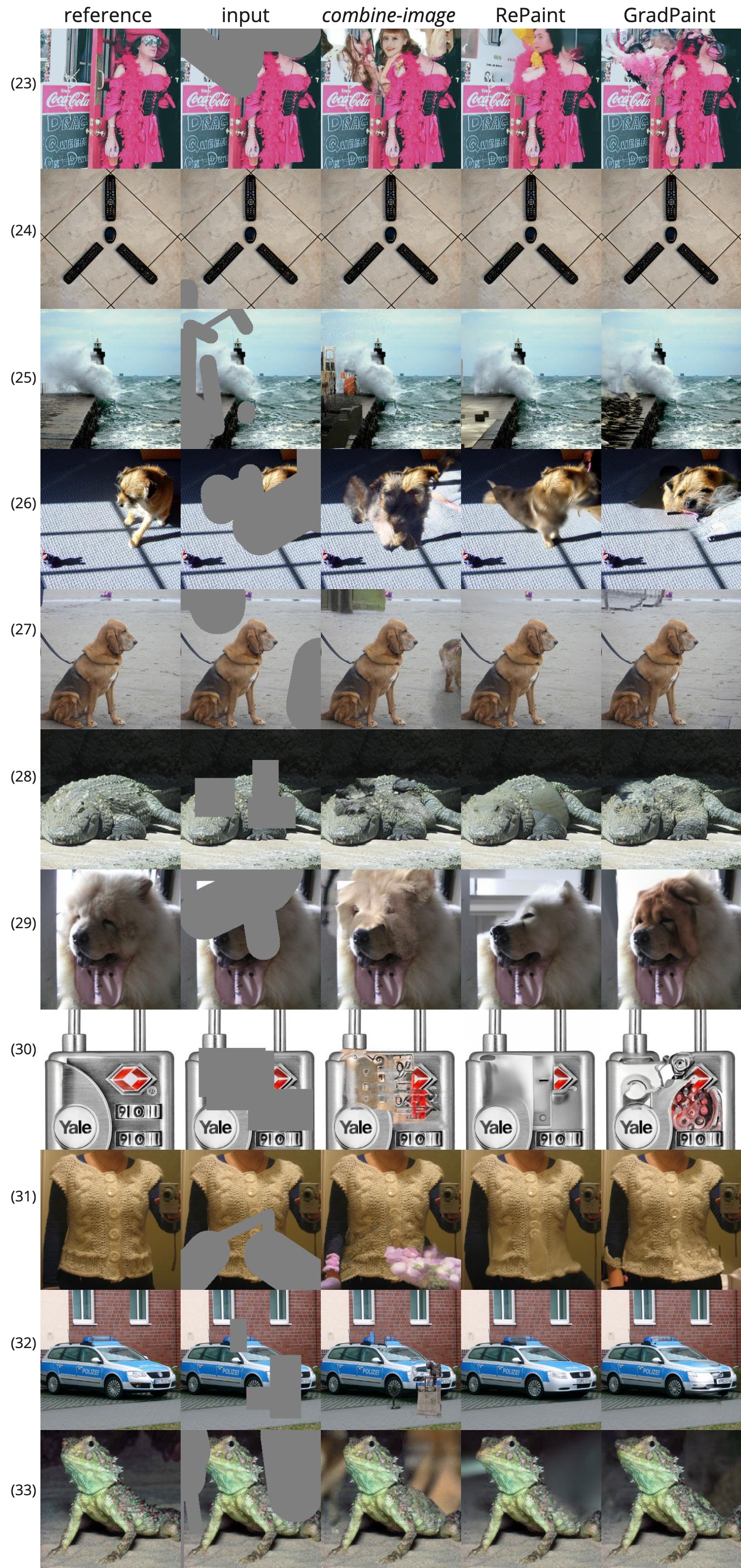}
    \caption{Uncurated results on ImageNet (3). Our method works well on global coherence (row 29) and generating fine textures requiring alignment (row 31).}
    \label{fig:uncur3}
\end{figure*}

\begin{figure*}[htbp]
  \centering
    \includegraphics[width=0.57\linewidth]{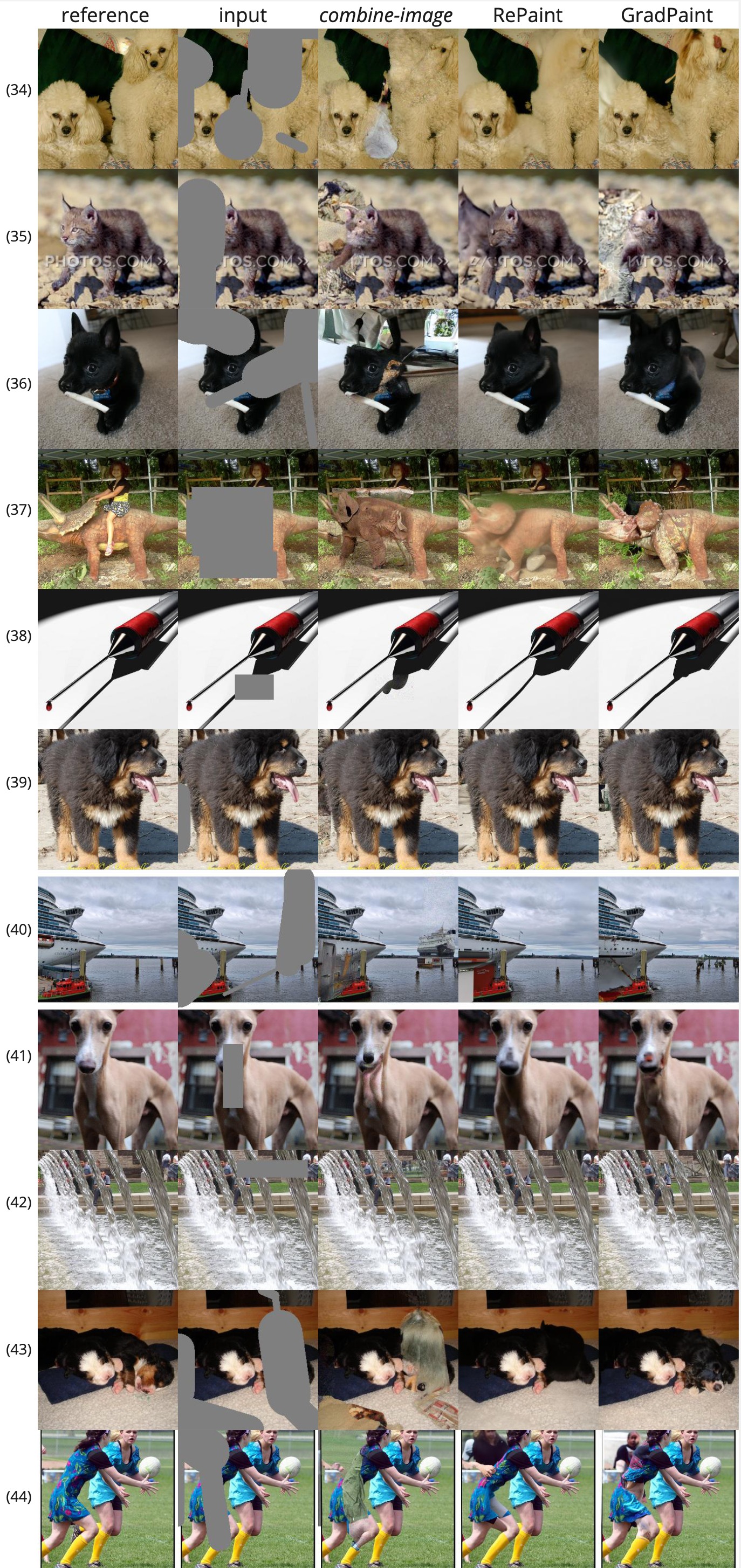}
    \caption{Uncurated results on ImageNet (4). Our method works well on global coherence (rows 36, 37) but sometimes fails on challenging tasks (row 35). From time to time, the background of the image may poorly influence the content to be generated (row 41).}
    \label{fig:uncur4}
\end{figure*}

We primarily previously showed select results on challenging tasks with masks in key places to better appreciate the differences between methods. Figs.~\ref{fig:uncur1}, ~\ref{fig:uncur2}, ~\ref{fig:uncur3}, and ~\ref{fig:uncur4} show uncurated examples from on ImageNet comparing \emph{combine-image}, RePaint, and GradPaint. GradPaint works particularly well when the task requires fine-grained texture alignment (see rows 1, 5, 12, 31)  and tends to work significantly better for global image coherence compared to other methods (see rows 2, 4, 6, 29, 36, 37). However, GradPaint sometimes fails with difficult tasks with unstandard context (see rows 14, 36). From time to time, we see that GradPaint may slightly add unneeded bias to the inpainting task from the background, e.g. row 41 produced a dog with a pink nose, influenced by the pink background. In general, GradPaint works better than RePaint, even though RePaint requires 5x the compute time (or 7x the compute time with the modified GradPaint presented in \ref{speed_section}).

\end{document}